\ifdefined\ACMReviewMode\else
  \def\ACMLengthCheckMode{1}
\fi
\ifdefined\ACMLengthCheckMode
  \documentclass[acmsmall]{acmart-arxiv}
\else
  \documentclass[manuscript,review]{acmart-arxiv}
\fi
\setcopyright{none}
\usepackage{longtable}
\usepackage{array}
\usepackage{booktabs}
\usepackage{colortbl}
\usepackage{calc}
\graphicspath{{figures/}}
\providecommand{\real}[1]{#1}
\definecolor{TableHeader}{gray}{0.92}
\definecolor{TableRule}{gray}{0.38}
\definecolor{TableRowRule}{gray}{0.78}
\AtBeginEnvironment{longtable}{\footnotesize\setlength{\tabcolsep}{3pt}\setlength{\extrarowheight}{0.25pt}\renewcommand{\arraystretch}{1.00}\arrayrulecolor{TableRule}}
\AtEndEnvironment{longtable}{\arrayrulecolor{black}}
\AtBeginDocument{\raggedbottom}
\newcommand{\TableTopRule}{\arrayrulecolor{TableRule}\toprule}
\newcommand{\TableMidRule}{\arrayrulecolor{TableRule}\midrule}
\newcommand{\TableBottomRule}{\arrayrulecolor{TableRule}\bottomrule\arrayrulecolor{black}}
\newcommand{\TableRowSep}{\\[-0.2ex]\arrayrulecolor{TableRowRule}\specialrule{0.25pt}{0.20ex}{0.20ex}\arrayrulecolor{TableRule}}
\ifdefined\ACMLengthCheckMode
\else
\fi
\acmJournal{CSUR}
\acmDOI{}
\acmISBN{}
\acmVolume{}
\acmNumber{}
\acmArticle{}
\acmYear{}
\acmMonth{0}
\copyrightyear{}
\makeatletter
\renewcommand{\@mkbibcitation}{\begingroup
  \let\@vspace\@vspace@orig
  \let\@vspacer\@vspacer@orig
  \par\medskip\small\noindent\textbf{ACM Reference Format:}\par\nobreak
  \noindent\begingroup\def\\{\unskip{}, \ignorespaces}\authors\endgroup.
  \@title. Unpublished manuscript prepared for submission to
  \textit{ACM Computing Surveys}, \ref{TotPages} pages.\par
\endgroup}
\makeatother
\ifdefined\ACMLengthCheckMode
  \renewcommand{\footnotetextcopyrightpermission}[1]{}
  \AtBeginDocument{%
    \fancyfoot{}%
    \fancyfoot[RO,LE]{\footnotesize Manuscript prepared for submission to ACM Computing Surveys}%
    \fancypagestyle{firstpagestyle}{%
      \fancyhf{}%
      \fancyfoot[RO,LE]{\footnotesize Manuscript prepared for submission to ACM Computing Surveys}}}
\fi
\title[Hallucination Across the Reasoning Lifecycle]{Hallucination Across the Reasoning Lifecycle: Interface Visibility, Causal Evidence, and Release Control in Large Reasoning Models}
\author{Zhe Yu}
\affiliation{%
  \institution{Zhejiang University}
  \city{Hangzhou}
  \country{China}}
\affiliation{%
  \institution{Binjiang Institute of Zhejiang University}
  \city{Hangzhou}
  \country{China}}
\email{zyu@zju-if.com}

\let\PreviewOriginalCity\city
\renewcommand{\city}[1]{%
  \PreviewOriginalCity{#1}%
  \ifstrequal{#1}{Hong Kong SAR}{, #1}{}\ignorespaces}
\author{Mohan Li}
\affiliation{%
  \institution{Guangzhou University}
  \city{Guangzhou}
  \country{China}}
\email{limohan@gzhu.edu.cn}

\author{Lei Yu}
\affiliation{%
  \institution{Rensselaer Polytechnic Institute}
  \city{Troy}
  \state{New York}
  \country{USA}}
\email{yul9@rpi.edu}

\author{Ka-Ho Chow}
\affiliation{%
  \institution{The University of Hong Kong}
  \city{Hong Kong SAR}
  \country{China}}
\email{kachow@cs.hku.hk}

\author{Chengwei Qin}
\affiliation{%
  \institution{The Hong Kong University of Science and Technology (Guangzhou)}
  \city{Guangzhou}
  \country{China}}
\email{chengweiqin@hkust-gz.edu.cn}

\author{Xingyu Wu}
\affiliation{%
  \institution{The Hong Kong Polytechnic University}
  \city{Hong Kong SAR}
  \country{China}}
\email{xingy.wu@polyu.edu.hk}

\author{Wenpeng Xing}
\affiliation{%
  \institution{Zhejiang University}
  \city{Hangzhou}
  \country{China}}
\email{wpxing@zju.edu.cn}

\author{Shuguang Xiong}
\affiliation{%
  \institution{Research Center for Scientific Data Hub, Zhejiang Lab}
  \city{Hangzhou}
  \country{China}}
\email{xiongshuguang@zhejianglab.org}

\author{Meng Han}
\affiliation{%
  \institution{Zhejiang University}
  \city{Hangzhou}
  \country{China}}
\email{mhan@zju.edu.cn}
\renewcommand{\shortauthors}{Yu et al.}

\begin{abstract}
Reasoning errors can propagate into later decisions and memory. This survey synthesizes 312 papers and first-party reports on text-based reasoning hallucinations around three questions: what evidence is observable, what study designs establish, and which corrective actions the evidence supports. UIPCA records unsupported premises (U), invalid inferences (I), dependent reuse (P), visible answer--trace consistency (C), and action-policy failures (A). Across 58 reviewed sources, no comparison establishes that a specified intervention improves reasoning while reducing factual reliability under matched conditions. The synthesis connects diagnosis to verification, repair, selective release, and persistent-state control across memory, tools, and training feedback.
\end{abstract}

\ccsdesc[500]{Computing methodologies~Natural language generation}
\ccsdesc[500]{Computing methodologies~Machine learning}
\keywords{Hallucination, Large Reasoning Models, Chain-of-Thought, Factuality, Faithfulness, Verification}
\begin{document}
\maketitle

\section{Introduction}\label{introduction}

The same wrong answer can arise from distinct failures. A reasoning model may start from an unsupported premise, make an invalid inference, reuse an earlier error, or contradict its own visible reasoning. Repair may require retrieving evidence, correcting a step, revisiting a dependent branch, checking answer--trace consistency, or abstaining. Final-answer accuracy alone cannot select the appropriate repair.

These errors can also affect later decisions. Agents may store generated content in private or shared memory, act through tools, or retain feedback for later rounds \citep{shinn2023reflexion,yao2023react,du2023multiagent}. An unsupported claim can then become future context. Storage alone does not establish propagation: the return path depends on whether the claim is reused before independent repair.

\begin{figure}[!t]
  \centering
  \scalebox{1}[0.88]{\includegraphics[width=\columnwidth]{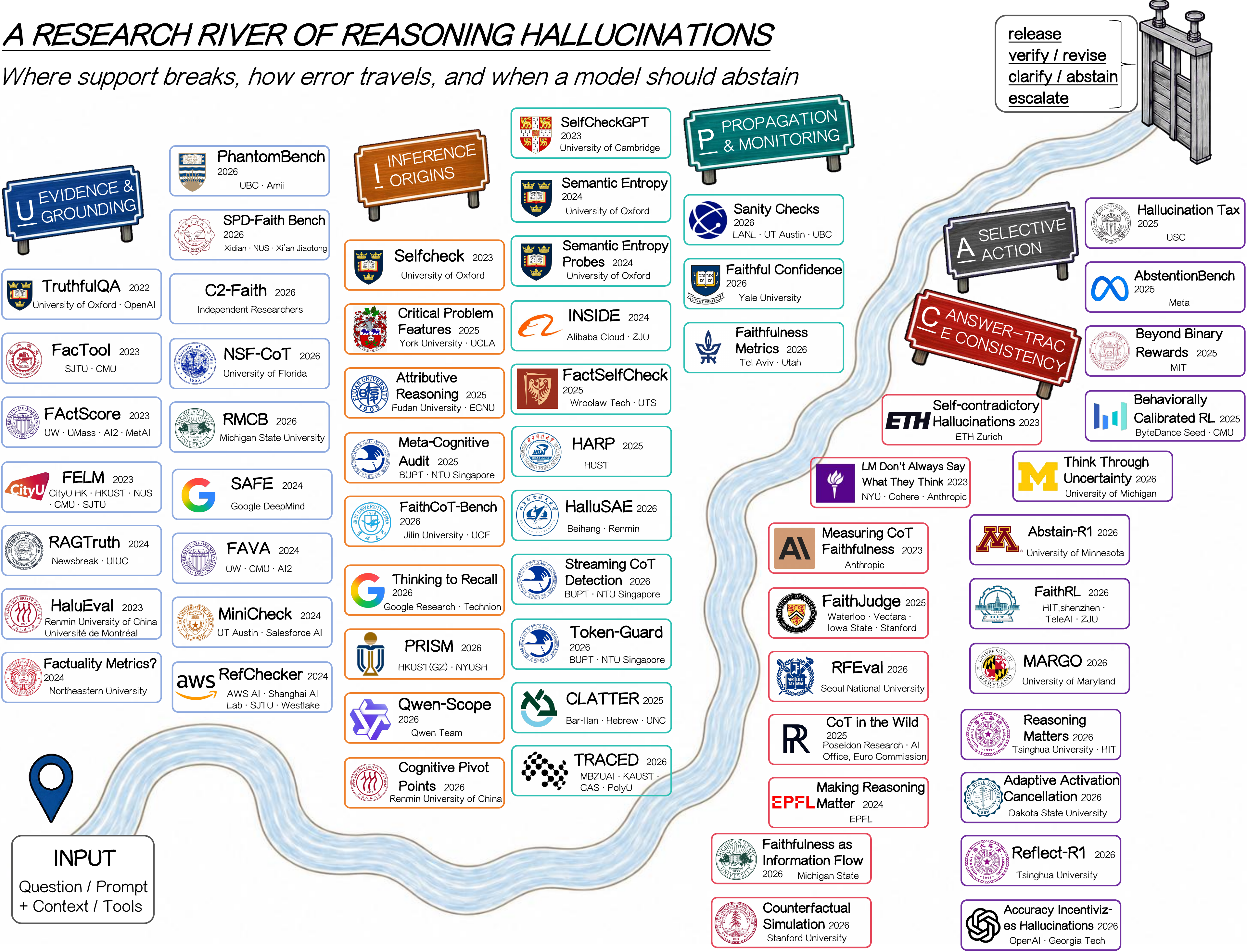}}
  \caption{Research river of reasoning hallucinations: 60 representative works grouped by primary analytical role. The regional headings name research themes, not formal UIPCA fields. In particular, ``Propagation \& Monitoring'' groups related literature; \(P\) records dependent reuse of a confirmed error, which repetition or a monitoring signal alone cannot establish. Likewise, the consistency region includes process-faithfulness work, while \(C\) checks only visible answer--trace consistency. The separate action region informs the \(A\) policy audit. Placement denotes neither chronology nor causal order, and groups are not exclusive. S7 provides canonical titles and links.}\label{fig:research-river}
  \Description{A winding river runs from an input marker to a release gate. Sixty paper cards with years and institutional marks are grouped under evidence and grounding, inference origins, propagation and monitoring, answer--trace consistency, and selective action. These headings organize research themes rather than define formal UIPCA fields. In particular, P requires dependent reuse of a confirmed error; C compares the answer with the visible trace. The separate action region informs policy auditing. The river is not a temporal or causal sequence.}
\end{figure}

Existing taxonomies, monitoring methods, and interventions address different parts of this problem. This survey connects three questions: what the evaluator can observe, what the study design establishes, and which corrective action the evidence supports. Their separation can obscure the difference between consistency and correctness, association and causation, or detection and a justified release decision. Figure~\ref{fig:research-river} maps the literature supporting this evidence-to-action view.

A correct answer can follow unsupported reasoning, while coherent deductions from a distorted graph, mismatched image region, or invented API can yield a wrong answer. Correction creates another distinction: accepting a local edit does not establish recovery of the final outcome \citep{al2025auditing}.

These failures arise in several kinds of reasoning system. Chain-of-thought prompting asks a model to generate intermediate steps; self-consistency samples several reasoning paths and aggregates their answers \citep{al2022chainofthought,al2022large,al2022leasttomost,al2022selfconsistency}. Reasoning post-training instead changes the model's parameters \citep{deepseekai2025deepseekr,al2025are}. Whatever the method, a provider may disclose a trace, a summary, or neither \citep{openai2024openai,anthropic2025claude,googledeepmind2025gemini,qwen2025team}. A fluent explanation is therefore something to check, not proof of how the answer was obtained.

These distinctions change how findings should be read. Factual QA can improve or decline across post-training recipes, but such comparisons change complete pipelines \citep{al2025are}. Within a fixed checkpoint, extra deliberation can improve recall even though traces containing a wrong fact have lower final accuracy \citep{gekhman2026recall}. The latter association does not isolate the effect of introducing that fact. Neither result establishes that a specified intervention improves reasoning capability while reducing factual reliability.

For consequential claims, the relevant evidence may be a source, image region, formal constraint, or executable test \citep{al2024large,al2023a,liao2026medheal,lee2025codehallucination}. A final-answer judge cannot recover evidence never exposed, and a detector score alone cannot select an appropriate action. Deployment therefore requires more than detection: it requires an available check and a policy for answering, verifying, revising, or abstaining.

\begin{figure}[!t]
  \centering
  \input{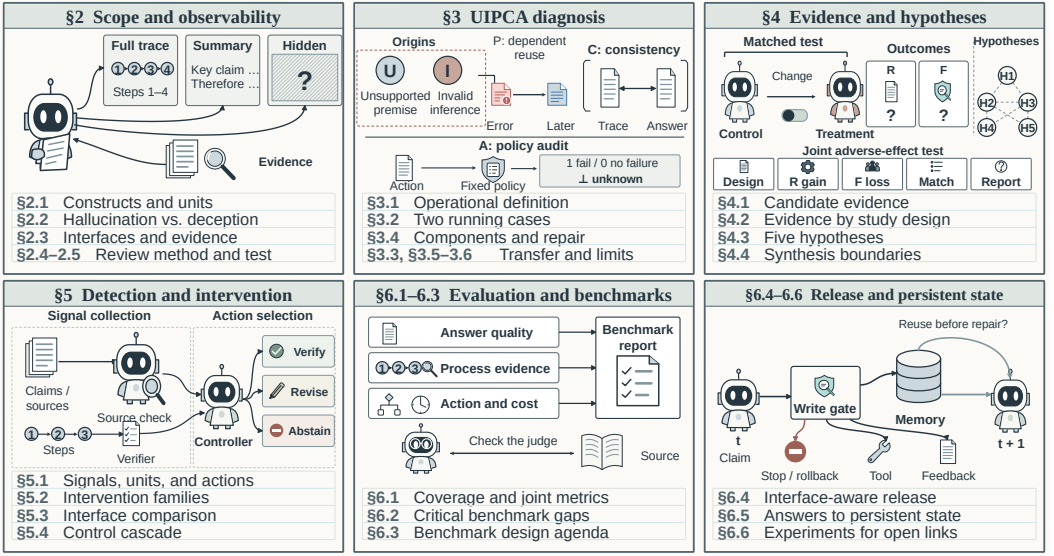}
  \caption{A guide to the review. Click section indices to open the six topics. In the evidence panel, \(R\) denotes reasoning capability and \(F\) factual reliability. The five checks concern the joint adverse-effect criterion, not study classes. Dashed links among H1--H5 are schematic, not causal claims.}\label{fig:survey-navigation}
  \Description{Six panels with section indices summarize the review. Diagrams show interface visibility, the UIPCA diagnostic record and separate policy audit, joint adverse-effect checks and hypotheses, detection signals and action selection, benchmark reporting, and checks before memory, tool, or feedback writes.}
\end{figure}

Existing surveys cover adjacent questions but use different organizing units. Broad hallucination reviews organize definitions, causes, benchmarks, detection, and mitigation across NLG and LLMs \citep{al2023survey,huang2025hallucinationsurvey,zhang2025sirenssong}. General and domain reviews extend that map to foundation models, multimodal failure, and code defects \citep{al2024comprehensive,alansari2025llm,cossio2025a,pesaranghader2026hallucination,lee2025codehallucination}. Reasoning surveys emphasize methods, training, evaluation, and trustworthiness \citep{mondorf2024multistep,al2025trustworthiness,anand2026the,hu2026mechanistic}.

Specialized reviews organize failures by model lifecycle \citep{ho2026understanding}, reasoning type \citep{song2026reasoningfailures}, trace length \citep{chen2026longcot}, or image-grounded training and task demand \citep{dong2026imagegroundedcot}. Liu et al.\ instead derive a taxonomy and test two interventions in an educational application \citep{liu2026reasoninghallucinationsurvey}. Table~\ref{tab:survey-positioning} compares their scope and organizing principles.

The review makes three contributions. First, UIPCA provides a common record of error origins, dependent reuse, visible consistency, and policy action. Second, a study-design audit tests the reasoning--reliability evidence against the matched joint adverse-effect criterion. Third, the intervention synthesis compares verification, repair, abstention, escalation, and write control across full-trace, summary, and hidden-reasoning interfaces.

The closest process taxonomy, \emph{Mirages of Logic}, separates premise, operation, logic, and conclusion hallucinations \citep{anonymous2026mirages}. UIPCA retains such local error types while recording dependent reuse, interface-specific unknowns, visible answer--trace consistency, and action-policy failure. These distinctions matter because the same local error may require different repairs depending on its downstream use, observability, and consequences. S4 compares the two schemes on the same cases.

\begin{longtable}[]{@{}
  >{\scriptsize\raggedright\arraybackslash}p{(\linewidth - 6\tabcolsep) * \real{0.2100}}
  >{\scriptsize\raggedright\arraybackslash}p{(\linewidth - 6\tabcolsep) * \real{0.2350}}
  >{\scriptsize\raggedright\arraybackslash}p{(\linewidth - 6\tabcolsep) * \real{0.2750}}
  >{\scriptsize\raggedright\arraybackslash}p{(\linewidth - 6\tabcolsep) * \real{0.2800}}@{}}
\caption{What neighboring surveys organize, and which joint gap this review addresses.}\label{tab:survey-positioning}\\
\TableTopRule\noalign{}
\rowcolor{TableHeader}\textbf{Survey / Focus} & \textbf{Interface and Evidence Design} & \textbf{Persistence and Action} & \textbf{What This Review Adds} \\
\TableMidRule\noalign{}
\endfirsthead
\caption[]{(continued)}\\
\TableTopRule\noalign{}
\rowcolor{TableHeader}\textbf{Survey / Focus} & \textbf{Interface and Evidence Design} & \textbf{Persistence and Action} & \textbf{What This Review Adds} \\
\TableMidRule\noalign{}
\endhead
\TableBottomRule\noalign{}
\endlastfoot
Broad and domain hallucination surveys \citep{al2023survey,huang2025hallucinationsurvey,zhang2025sirenssong,al2024comprehensive,lee2025codehallucination} & Organize definitions, causes, benchmarks, mitigation, or domain-specific errors; interface disclosure is not the main axis & Controls are usually attached to an output, model stage, or domain failure & Retain their error types and checks, then record visibility, evidence design, dependency reuse, and action \TableRowSep
Ho et al.\ \citep{ho2026understanding} & Multi-level definition and model lifecycle connect text and vision; causal matching is not the main organizing idea & Causes and mitigations span model families, but released claims are not tracked through dependencies or later writes & Add interface-specific unknown labels, study-design classes, and claim reuse \TableRowSep
Song et al.\ \citep{song2026reasoningfailures} & Crosses reasoning types with fundamental, application, and robustness failures & Identifies causes and mitigation directions across informal, formal, and embodied reasoning & Add error location and separate process diagnosis from release-policy failure \TableRowSep
Long- and image-grounded CoT surveys \citep{chen2026longcot,dong2026imagegroundedcot} & Compare trace length, training method, task demand, or interleaved states & Emphasize capability and method trade-offs rather than later unsupported writes & Use their method distinctions while checking support, visibility, and action \TableRowSep
Liu et al.\ \citep{liu2026reasoninghallucinationsurvey} & Evaluates reasoning hallucination in an educational setting and derives a task-specific classification & Tests uncertainty-based abstention and model-to-model correction within that application & Supply the broader, source-level evidence synthesis and interface-aware diagnostic record \TableRowSep
\emph{Mirages of Logic} \citep{anonymous2026mirages} & Types visible premise, operation, logic, and conclusion failures & Supports local process diagnosis; dependency reuse, interface opacity, and release action are outside its core type labels & Retain error type, then add origin/reuse, unknown-by-interface, answer--trace consistency, and policy action \\
\end{longtable}

\begin{figure}[t]
  \centering
  \includegraphics[width=\columnwidth]{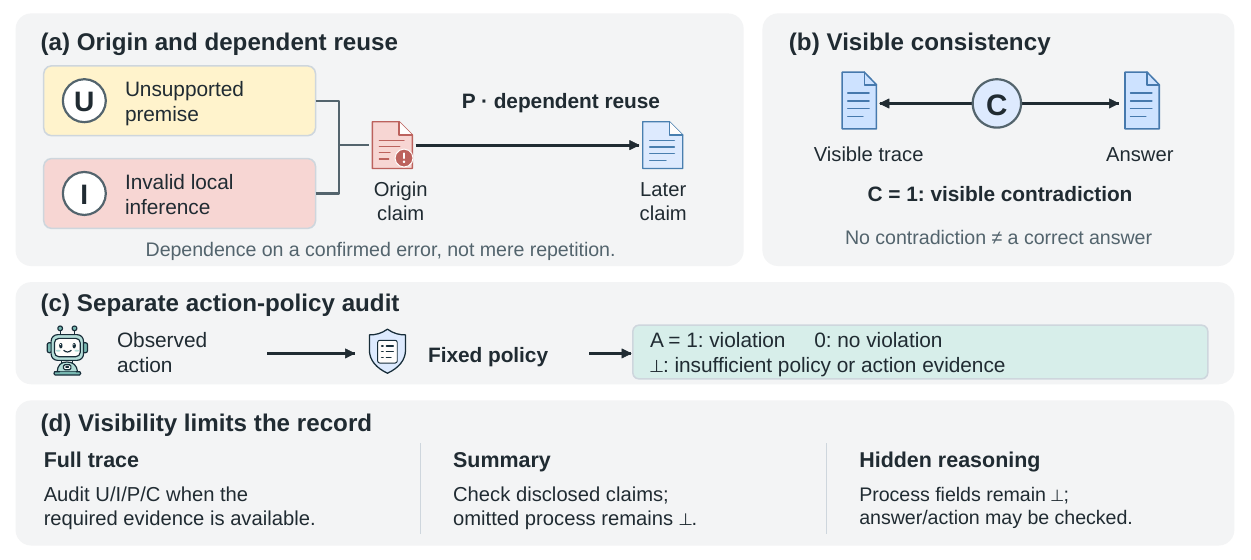}
  \caption{UIPCA diagnostic record. Panel (a) locates unsupported premises or invalid inferences (\(U/I\)) and dependent reuse (\(P\)). Panel (b) compares the answer with the visible trace (\(C\)), not hidden computation. Panel (c) audits the observed action against a fixed policy (\(A\)). Panel (d) shows interface limits: unavailable evidence remains \(\bot\), meaning unknown. The fields answer different diagnostic questions, not five sequential stages.}\label{fig:taxonomy}
  \Description{Four panels separate origin and dependent reuse, visible trace--answer consistency, action-policy compliance, and interface limits. U and I are alternative origin labels. P connects an origin claim to a dependent later claim. C compares trace and answer. A is a separate policy audit. Full, summary, and hidden interfaces permit different checks.}
\end{figure}

Figure~\ref{fig:taxonomy} separates the diagnostic fields, and Figure~\ref{fig:deployment-workflow} extends them to retained outputs.

The review asks how to diagnose and repair failures (\textbf{RQ1}) and what current evidence establishes about reasoning and factual reliability (\textbf{RQ2}). It then examines which detection and intervention strategies different interfaces permit (\textbf{RQ3}), and how benchmarks and release policies should evaluate consequential outputs and persistent state (\textbf{RQ4}). Sections~\ref{scope-evidence-method}--\ref{evaluation-deployment-agenda} address these questions in turn (Figure~\ref{fig:survey-navigation}). S1--S7 provide the corpus, evidence reviews, coding rules, catalogs, and figure data.

\textbf{Scope.} This is a text-centered review of reasoning hallucination. Multimodal and code studies are transfer cases, not co-equal domain surveys. They test which distinctions survive a change in evidence type and where a domain-specific checker is required.

\section{Scope, Evidence Model, and Review Method}\label{scope-evidence-method}

\subsection{Constructs and Evaluation Units}\label{constructs-interfaces}

Here, \emph{large reasoning model} means a system that allocates or supervises multi-step computation beyond direct generation. This includes visible scratchpads, hidden effort modes, sampled search, tool use, and reasoning-focused post-training. A displayed rationale is generated text, not necessarily a transcript of hidden computation.

Training methods shape what these systems learn to produce. After pretraining \citep{al2023llama}, supervised fine-tuning (SFT) learns from demonstrations, while reinforcement learning (RL) rewards selected behavior. Feedback can concern a final answer or individual reasoning steps, and a verifier checks proposed answers or steps \citep{al2022instructgpt,al2021training,al2023lets,al2024deepseekmath}. Distillation trains a model using another model's outputs. In practice, data, rewards, optimization, and the retention of reasoning steps often change together \citep{al2025dapo,al2025openreasonerzero,al2025understanding,li2025causal}. Labels such as RL-only or SFT+RL describe a training recipe, not a single isolated cause.

In this review, hallucination means generated content that lacks support under the task's stated reference frame. Two familiar checks capture different aspects of support: \emph{factuality} asks whether a claim is correct, while \emph{source faithfulness} asks whether the supplied evidence supports it \citep{al2020on,al2023survey,al2023factscore}. For example, a true statement drawn from outside a designated document may still be unsupported in a document-grounded answer.

\emph{Process faithfulness} concerns the explanation rather than the source: does the displayed reasoning reflect what actually determined the answer \citep{al2023measuring}? A coherent explanation can fail this test if the model reached its answer for another reason. Checking the explanation's statements and testing its causal role are therefore separate tasks.

Confidence and correction add two further questions. A calibrated system's stated confidence agrees with its observed correctness; corrigibility concerns whether valid feedback repairs later reasoning and the outcome \citep{al2025auditing}. Neither guarantees a good decision to act. A system may recognize uncertainty yet release an answer that should have been verified or withheld. Table~\ref{tab:construct-interface} separates these evaluation targets so that success on one is not mistaken for success on all.

\subsection{Boundary Between Hallucination and Deception}\label{hallucination-deception-boundary}

Hallucination and deception can produce the same false statement, but deception requires evidence of goal-directed concealment or strategic misrepresentation \citep{hagendorff2024deception,al2024from}. A false, persistent, or self-contradictory answer does not establish intent. Incentive changes and tests of information access are needed to support a strategic explanation \citep{al2024sleeper,al2025alignmentfaking}.

UIPCA records observable support, dependency, consistency, and action failures without inferring mental state. The same record can arise from missing knowledge, invalid inference, reward pressure, or concealment. A factually supported answer may have no \(U/I/P/C\) failure and still violate a declared release policy; \(A=1\) records that policy failure.

\subsection{Observable Interfaces and Evidence Status}\label{observable-interfaces}

The three interfaces differ in what the evaluator can inspect. A full trace exposes ordered reasoning steps; a summary discloses selected claims; a hidden interface exposes no reasoning record. In our comparison, the prompt, designated evidence, answer, and observed action remain available. A summary can assist answer checking, especially when the evaluator lacks the prompt \citep{algaba2026observability}, but it cannot establish what happened in omitted steps.

Visibility can also change behavior. Disclosure affects abstention, and length-penalized training can remove monitoring cues with little accuracy loss \citep{mendi2026why2speak,little2026lengthpenalties}. Viewing the same run through different interfaces is thus distinct from changing how a model generates or compresses its reasoning.

A missing source has different meanings in different tasks. If an answer must be supported by a supplied document, absence of the required support establishes a grounding failure. In an open-world factual search, failing to find a source does not establish that the claim is false. We therefore distinguish supported, unsupported, and unknown claims. Unknown means that the available evidence cannot decide; it is not a favorable judgment. S4 gives the formal rules.
\begingroup\setlength{\LTpre}{0.65\baselineskip}\setlength{\LTpost}{0.65\baselineskip}
\begin{longtable}[]{@{}
  >{\scriptsize\raggedright\arraybackslash}p{(\linewidth - 6\tabcolsep) * \real{0.1800}}
  >{\scriptsize\raggedright\arraybackslash}p{(\linewidth - 6\tabcolsep) * \real{0.2850}}
  >{\scriptsize\raggedright\arraybackslash}p{(\linewidth - 6\tabcolsep) * \real{0.2550}}
  >{\scriptsize\raggedright\arraybackslash}p{(\linewidth - 6\tabcolsep) * \real{0.2800}}@{}}
\caption{What each evaluation target measures and what the interface permits an evaluator to establish.}\label{tab:construct-interface}\\
\TableTopRule\noalign{}
\rowcolor{TableHeader}\textbf{Evaluation Target} & \textbf{Operational Question} & \textbf{Typical Unit / Evidence} & \textbf{Interface Boundary} \\
\TableMidRule\noalign{}
\endfirsthead
\caption[]{(continued)}\\
\TableTopRule\noalign{}
\rowcolor{TableHeader}\textbf{Evaluation Target} & \textbf{Operational Question} & \textbf{Typical Unit / Evidence} & \textbf{Interface Boundary} \\
\TableMidRule\noalign{}
\endhead
\TableBottomRule\noalign{}
\endlastfoot
Factuality / source grounding & Is a claim true at evaluation time, or licensed by the designated source? & Atomic claim or answer; fact label, source entailment, execution result & Available under all interfaces when the required external evidence exists \TableRowSep
Process faithfulness & Does the displayed reasoning reflect what causally determines the answer? & Trace--answer pairs with controlled perturbations \citep{al2023measuring,hashemi2025measuring} & Causal attribution requires intervention; textual consistency alone is insufficient \TableRowSep
Uncertainty calibration & Do reported probabilities or scores match empirical correctness? & Answer set; Brier, ECE, calibration curve \citep{al2025rlcr,al2026rmcb} & Behavioral confidence can be evaluated under all interfaces; internal signals require provider access \TableRowSep
Corrigibility & Does valid feedback change the dependent trace and outcome? & Edited step or trajectory \citep{al2025auditing} & Requires before/after evidence; local acceptance is not outcome recovery \TableRowSep
Selective decision quality & Does a fixed policy choose answer, retrieve, verify, clarify, abstain, or escalate appropriately? & Item and action; selective risk, coverage, action counts & The action is observable; its justification depends on declared policy inputs and evidence \\
\end{longtable}
\endgroup
The lifecycle is a separate diagnostic layer, not another evaluation target or model capability. \(U/I\) locate unsupported premises or invalid inferences, \(P/C\) describe reuse and consistency, and \(A\) records a violation of a policy fixed in advance. These fields help interpret the targets in Table~\ref{tab:construct-interface}; they do not replace their measures.

\subsection{Search, Selection, and Evidence Strength}\label{review-methodology}

This review draws on 312 papers and first-party reports available by 30 August 2026. We combined OpenAlex and arXiv searches with citation tracing and targeted checks of underrepresented topics. Selection focused on reasoning hallucination, its detection and control, and benchmarks that expose relevant failures. Multimodal and code studies test transfer of the framework; they do not form equally comprehensive subreviews.

\begin{figure}[!t]
  \centering
  \includegraphics[width=\columnwidth]{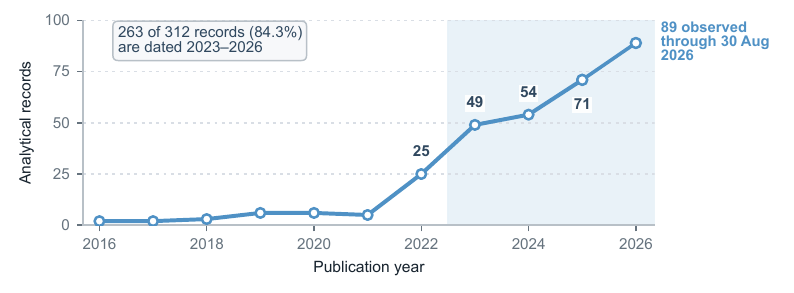}
  \caption{Publication-year distribution of the 312-record analytical corpus. The solid blue series reaches 89 records observed through 30 August 2026; 263 records (84.3\%) are dated 2023--2026. The 2026 count covers only part of the year. These counts describe the curated collection, not exhaustive field-wide publication totals.}\label{fig:corpus-trend}
  \Description{A solid blue line charts annual analytical-corpus records from 2016 through the partial 2026 observation. Counts remain below ten through 2021, rise to 25 in 2022, then to 49, 54, and 71 in 2023 through 2025. The solid blue segment reaches 89 observed records through 30 August 2026. A callout states that 263 of 312 records (84.3 percent) are dated 2023 through 2026. No full-year estimate is shown.}
\end{figure}

The collection is curated rather than an exhaustive census of the field. Main-text sources support the synthesis and illustrate consequential differences between methods; the complete catalog retains additional coverage. Figure~\ref{fig:corpus-trend} shows publication years within this collection. Exact queries, eligibility and exclusion rules, screening decisions, and the separate review of reasoning--reliability comparisons are available in S1--S3. The accompanying \emph{Supplementary Methods} PDF explains these procedures without requiring readers to inspect the data files.

For the reasoning--reliability question, relevance screening of all 312 records retained the 20 original S3 sources and identified 14 additional candidates for criterion-level review. Targeted searches and citation tracing from S3 sources sought further relevant comparisons; title-based prioritization yielded 24 additional sources with retrievable primary text for review. These 34 corpus sources and 24 external sources form the criterion-review set, not a second count of the analytical corpus (S3).

Study design determines how these sources inform the argument. Changing one factor in a controlled experiment can test its effect. Comparing complete systems instead reveals a combined outcome, while grouping naturally generated traces shows association. Section~\ref{evidence-by-design} uses this distinction to explain findings that might otherwise appear contradictory.

Evidence grades distinguish \textbf{direct, single-setting evidence} for a stated link, \textbf{supported synthesis} across compatible findings, and \textbf{limited or mixed evidence} that is sparse or conflicting. A \textbf{proposed test} has no result yet; other cases remain \textbf{unresolved}.

\textbf{Use of automated and language-model tools.} Deterministic scripts applied the frozen screening rules and generated release ledgers. Language models assisted with sentence-level drafting and prose revision. The authors defined the protocol, selected the analytical sources, verified the evidence used for manuscript claims, determined the analysis, and take responsibility for the submitted text.

\subsection{Prospective Matched Joint Adverse-Effect Test}\label{formal-scope-notation}

Does improving reasoning come at the expense of factual accuracy? A stronger reasoning model with more factual errors may suggest this trade-off, but the comparison can also reflect different data, rewards, or evaluation tasks. Our \emph{matched joint adverse-effect test} asks a narrower question: does one specified intervention improve reasoning capability while reducing factual reliability?

For comparison \(j\), let \(\Delta R_j\) and \(\Delta F_j\) denote intervention-minus-control changes in reasoning capability and factual reliability, both measured on higher-is-better scales. The joint adverse effect is
\begin{equation}\label{eq:strict-divergence-criterion}
\mathcal D_j=\mathbf{1}\!\left[\Delta R_j>\delta_R\ \land\ \Delta F_j<-\delta_F\right],
\qquad \delta_R,\delta_F\geq 0.
\end{equation}
The margins \(\delta_R\) and \(\delta_F\) specify the smallest meaningful gain and loss, chosen before evaluation. Equation~\ref{eq:strict-divergence-criterion} defines the effect, not its statistical support. To establish it at a prespecified error level \(\alpha\), a joint \(1-\alpha\) confidence region must lie entirely above \(\delta_R\) and below \(-\delta_F\). Multiplicity-adjusted one-sided intervals can enforce the same requirement.

The comparison must isolate the stated treatment: inference-time experiments hold the checkpoint and non-target settings fixed; training experiments start from a common base and control non-target training choices. Both use matched evaluation items and evaluators. Neither trace length nor a ``reasoning model'' label substitutes for measured capability. Failure to meet the criterion leaves the trade-off unresolved; it does not establish absence of harm. S3 provides implementation and sensitivity rules. We next turn from comparisons between systems to diagnosis within a run.

\section{The Evidence-to-Action Lifecycle}\label{lifecycle-taxonomy}

UIPCA synthesizes diagnostic distinctions from the reviewed literature into a common record; its five fields do not describe an underlying model mechanism.

UIPCA answers a repair question: once a run is suspect, which observable check can still improve the decision? The \(U\) and \(I\) fields locate unsupported premises and invalid local inferences; \(P\) records dependent reuse. The \(C\) field compares the answer with the visible trace, while \(A\) audits the observed action against a fixed policy. The fields therefore serve different diagnostic purposes. UIPCA is neither a flat error taxonomy nor a claim that every run passes through five observable stages. Factual, logical, arithmetic, perceptual, attributional, and executable errors remain separate type labels.

\subsection{Operational Definition}\label{taxonomy-definition}

Consider a calculation based on a retrieved value. If the value is unsupported, the repair starts with its source (\(U\)); if the value is correct but the arithmetic is wrong, the repair starts with the inference (\(I\)). When later calculations reuse the error, correcting only its first occurrence leaves dependent results to repair (\(P\)). An error mentioned and then discarded does not create this propagation path.

The final answer adds two different checks. Does it agree with the visible reasoning (\(C\))? Was releasing it allowed by the policy fixed for the task (\(A\))? A wrong answer may agree with a wrong calculation, so consistency alone cannot establish correctness. Conversely, even a correct answer can violate a requirement for independent verification before release. These distinctions connect each diagnosis to a different repair.

In the figures and records, \(1\) means that the named failure is confirmed, \(0\) that it is ruled out, and \(\bot\) that evidence is insufficient. The first-error marker \(\tau^\star\) locates the earliest confirmed premise or inference failure. It need not locate the actual first error if earlier checks are unresolved. S4 contains the complete coding manual, aggregation rules, and boundary cases.

Visibility limits this record. With hidden reasoning, an evaluator can still check the answer against external evidence and inspect recorded tool actions, but cannot locate errors in unexposed steps. A summary supports checks on what it discloses, not reconstruction of what it omits. Missing process evidence remains unknown, rather than being inferred from confidence or agreement.

UIPCA is a diagnostic record, not an automatic detector. Human judgments, executable checks, and learned detectors can supply its evidence; Section~\ref{detection-intervention} compares their strengths and errors. The running cases below show what the record adds to answer scoring, while Section~\ref{annotation-limits} explains its sensitivity to analytical choices.

\subsection{Two Running Cases}\label{running-trace-cases}

\begin{figure}[t]
  \centering
  \includegraphics[width=\columnwidth]{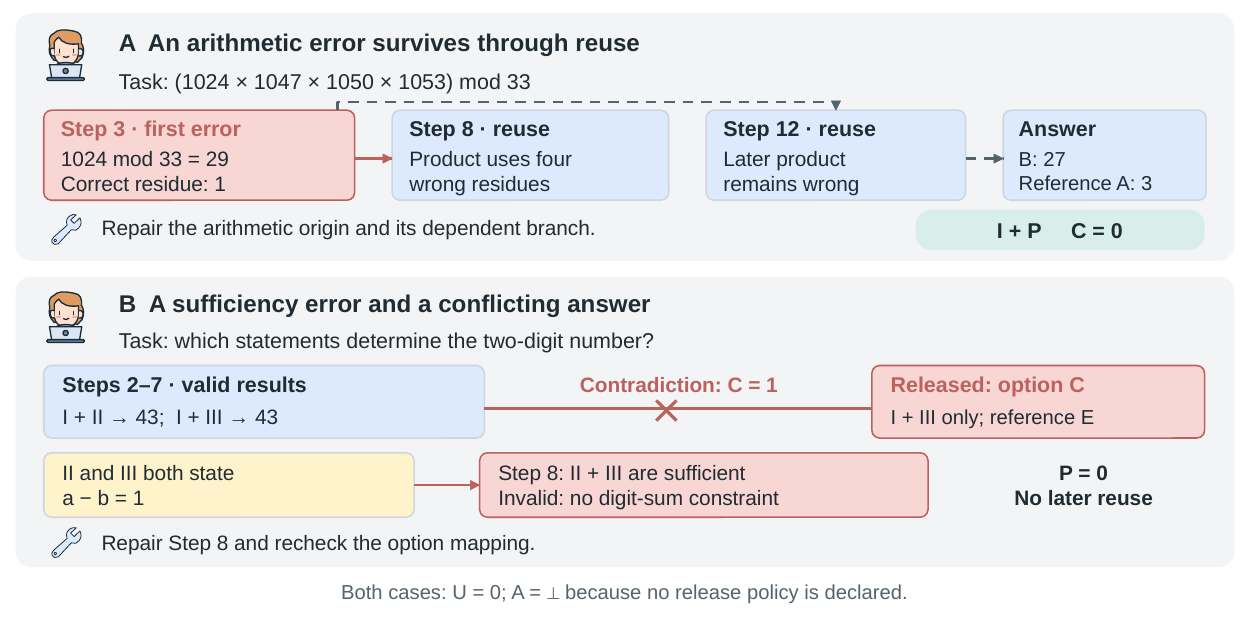}
  \caption{Two held-out FINE-CoT cases coded with U/I/P/C/A \citep{al2025faithcotbench}. Case A reuses an invalid Step 3 reduction (\(I+P\)); its answer follows the wrong trace, so \(C=0\). Dashed arrows abbreviate paths through omitted steps. Case B contains a separate Step 8 sufficiency error and a released option that conflicts with valid earlier results (\(I+C\)). The crossed line marks this contradiction, not dependence on Step 8. With no later reuse, \(P=0\).}\label{fig:trace-cases}
  \Description{Case A shows the task, an invalid Step 3 residue, reuse at Steps 8 and 12, and a wrong but trace-consistent answer. Case B separates valid earlier results from an invalid Step 8 sufficiency judgment. The released option conflicts with those earlier results; no dependency from Step 8 to the answer is asserted. Both cases keep the action label unknown because no release policy is declared.}
\end{figure}

In Case A, the trace assigns 29 instead of 1 to \(1024\bmod 33\), introduces three more wrong residues, and reuses them in later products \citep{al2025faithcotbench}. The initial arithmetic error is an \(I\) failure; later dependent calculations exhibit \(P\). The answer agrees with the wrong trace, so the consistency check finds no contradiction (\(C=0\)). Correcting the final number alone would leave the faulty derivation intact.

Case B requires a different repair. Steps 2--7 show that I+II and I+III each determine 43. Step 8 wrongly treats II+III as sufficient even though both state only \(a-b=1\) and give no digit sum \citep{al2025faithcotbench}. This is an inference error without later reuse. Separately, the released option C contradicts the valid earlier results, giving a consistency failure (\(C=1\)). Reference option E establishes that the answer is also wrong. The option comparison matters independently of Step 8.

Answer-only scoring marks both cases wrong but cannot select the repair. Case A needs correction before dependent reuse; Case B needs a constraint and option-mapping check despite no reuse of its Step 8 error. These are worked diagnoses, not estimates of how often either failure occurs. S4 preserves the source traces and reproducible coding checks.

\subsection{Text-Centered Evidence and Cross-Domain Transfer Tests}\label{cross-domain-lifecycle}

Selected multimodal and code studies test whether diagnostic distinctions survive a change of evidence; they do not estimate domain-wide prevalence or method rankings.

The checks differ: text requires sources, visual claims require objects or regions, and code permits execution without proving every requirement (Table~\ref{tab:cross-domain-lifecycle}).

\begin{longtable}[]{@{}
  >{\scriptsize\raggedright\arraybackslash}p{(\linewidth - 8\tabcolsep) * \real{0.1300}}
  >{\scriptsize\raggedright\arraybackslash}p{(\linewidth - 8\tabcolsep) * \real{0.2150}}
  >{\scriptsize\raggedright\arraybackslash}p{(\linewidth - 8\tabcolsep) * \real{0.2250}}
  >{\scriptsize\raggedright\arraybackslash}p{(\linewidth - 8\tabcolsep) * \real{0.2150}}
  >{\scriptsize\raggedright\arraybackslash}p{(\linewidth - 8\tabcolsep) * \real{0.2150}}@{}}
\caption{Shared lifecycle with domain-specific evidence and release consequences.}\label{tab:cross-domain-lifecycle}\\
\TableTopRule\noalign{}
\rowcolor{TableHeader}\textbf{Setting} & \textbf{Premise Object} & \textbf{Early Check} & \textbf{Later Check / Action} & \textbf{Unresolved Boundary} \\
\TableMidRule\noalign{}
\endfirsthead
\caption[]{(continued)}\\
\TableTopRule\noalign{}
\rowcolor{TableHeader}\textbf{Setting} & \textbf{Premise Object} & \textbf{Early Check} & \textbf{Later Check / Action} & \textbf{Unresolved Boundary} \\
\TableMidRule\noalign{}
\endhead
\TableBottomRule\noalign{}
\endlastfoot
Source-grounded text & Passage, database record, or time-indexed fact & Claim extraction and source entailment \citep{al2023factool,hu2024refchecker} & Verify, retrieve, revise, abstain & Hidden evidence use and open-world completeness \TableRowSep
Image or video & Object, region binding, frame, or temporal segment & Coverage and grounding \citep{al2023pope,al2024hallusionbench,2026spdfaith,ma2026hkvlm} & Reasoning check, targeted retrieval, abstention & Whether the rationale causally used the grounded evidence \TableRowSep
Function or repository code & Requirement, API, dependency, program state, repository snapshot & Type/contract/static check \citep{tian2024codehalu,erfanian2026delulu} & Sandbox, properties, hidden tests, patch review \citep{liu2023evalplus,jimenez2024swebench} & Untested paths, security, and unwritten intent \\
\end{longtable}

Multimodal reasoning makes the evidence boundary explicit. Caption and visual-QA benchmarks detect object hallucinations or image--answer disagreement \citep{al2018object,al2023faithscore,al2023openchair}, but do not by themselves distinguish a missed object from an ignored one. Visual-anchor and SPD-Faith analyses examine this distinction \citep{2026cognitive,2026spdfaith}. Region-binding and temporal-retrieval methods supply or check visual evidence for subsequent inferences \citep{ma2026hkvlm,chen2026reflectr1}. Medical and chemistry audits add a further warning: a correct answer does not make its displayed rationale valid evidence \citep{xu2026decorative,li2026chemicalcot}. Text-only deliberation cannot validate a missed visual detail without renewed access to the image, and answer scoring alone cannot identify the evidence used.

Executable code provides stronger local checks, but no test is universal. Execution exposes mapping, naming, resource, and logic errors. Program-state tasks isolate input--output reasoning, while containers catch invented symbols or imports \citep{tian2024codehalu,gu2024cruxeval,erfanian2026delulu}. Span-level detection across code, tool output, and documents shows why the evidence object must be recorded \citep{kovacs2026spangrounding}. Even ``tests pass'' is conditional evidence. EvalPlus rejects programs accepted by weaker suites and can change rankings \citep{liu2023evalplus}. Repository tasks also depend on issue interpretation, file selection, build state, and patch interaction \citep{jimenez2024swebench}. A code audit must preserve the environment and record any requirements that lack executable tests.

\subsection{Lifecycle Components and Their Repair Points}\label{lifecycle-loci}

\textbf{Premise entry (\(U\)).} A premise may be absent, stale, unsupported, or retrieved incorrectly. Short-answer benchmarks show the outcome \citep{wei2024simpleqa,al2017triviaqa}; cloze probes, knowledge-neuron analyses, retrieval comparisons, and bounded-corpus audits test whether needed information is recoverable \citep{al2019language,al2021knowledge,al2023when,al2025auditing}. These checks can motivate retrieval or source correction without revealing everything learned in pretraining. Image and code analogues include wrong regions and invented APIs.

\textbf{Local inference (\(I\)).} Proof audits and program-state tasks reveal invalid transitions despite correct premises \citep{al2025math,gu2024cruxeval}. But graph-coloring errors may originate in invented edges, followed by coherent deductions from the wrong graph \citep{heyman2025reasoning}. The distinction selects input repair versus rule correction. Structured methods separate premise selection and inference \citep{creswell2022faithful}; hybrids add symbolic solvers or rule checks \citep{pan2023logiclm,xu2024symbcot,pramanik2026nsfcot}.

\textbf{Propagation (\(P\)).} Later claims must depend on a confirmed error; length or repetition alone is insufficient \citep{al2025are,al2025auditing}. Independent evidence can stop reuse, whereas self-checks can repeat a shared false premise \citep{al2023selfcheckgpt,al2023selfcheck,al2024selfcontradictory}. In multi-round clinical evaluation, locally coherent rationales sometimes omitted decisive evidence requests before commitment \citep{braitsch2026informationseeking}. Repair should therefore examine both acquisition and reuse.

\textbf{Answer--trace consistency (\(C\)).} A trace can influence an answer or rationalize it, as cue, mediation, perturbation, and decomposition studies show \citep{al2023language,chua2025deepseekfaithful,al2025cot,cornish2025examining,paul2024making,radhakrishnan2023question}. Behavioral detection is easier for correct than wrong answers \citep{angdembay2026tworegimes}; causal-alignment tests instead intervene on trace dependence \citep{wang2026case}. The \(C\) field covers only visible contradiction, not causal attribution or hidden computation.

\textbf{Action (\(A\)).} Abstention and knowledge-boundary benchmarks reveal unjustified answers and refusals \citep{kirichenko2025abstentionbench,jung2026phantombench,al2025verify}. Reflection training and act-or-abstain tests connect explanations with decisions \citep{huang2026rein,mendi2026why2speak}, but verbal confidence alone is not a policy \citep{al2022language,al2022teaching,al2023can,al2025reasoning}. Audit false answers, unnecessary refusals, missed clarifications, and escalations against their costs and evidence requirements.

\subsection{Applying the Lifecycle Without Overclaiming}\label{applying-lifecycle}

A displayed dependency (\(P\)) or contradiction (\(C\)) need not describe hidden computation. Causal attribution requires changing a premise or step and observing the answer \citep{al2023measuring,hashemi2025measuring}. Textual diagnosis remains useful without treating rationales as mechanism transcripts.

The useful unit depends on the repair. A sentence may be appropriate for checking a source, yet contain several arithmetic operations that need separate checks. A token-level alarm can be precise without identifying which dependent branch should be restarted. Finer localization is useful only when it changes what can be repaired.

U/I/P/C/A records failure locations and relations, not severity. An unused unsupported premise can be serious, while a long failed branch can be harmless if discarded before release. Consequence and repair difficulty therefore remain separate from the number of recorded failures.

\subsection{Operational Sensitivity: Segmentation, Dependencies, and Unknowns}\label{annotation-limits}

A diagnostic record depends on how the trace is divided and which claims are treated as prerequisites. Merging a correct operation with a faulty one broadens the reported first-error location. Treating every neighboring step as dependent can also turn an isolated error into apparent propagation. These are changes in the analysis, not necessarily changes in the model's behavior.

Missing information creates a further limit. A trace can omit a prerequisite or present several derivations for one claim. In such cases, not finding a dependency is not enough to rule out reuse. The affected field remains unknown when the available evidence cannot decide. This is why a complete negative check and an unobserved check have different meanings.

Two laboratory members independently annotated 48 full-trace cases, blind to one another's labels and upstream evaluation labels. With 334 shared source steps, pre-adjudication Cohen's \(\kappa\) was 0.653 for \(U\), 0.730 for \(I\), and 0.858 for \(P\). Case-level \(C\) agreement was complete (\(\kappa=1.000\); four jointly positive cases). First-error location agreed in 41/48 cases; requiring the same origin type as well gave 40/48, including joint no-origin judgments. All \(A\) labels were unknown because no action policy was supplied. These estimates describe label agreement with fixed segmentation, not validation of the candidate dependency graph. S4 reports case-clustered intervals, positive/negative agreement, disagreement categories, and the status of edge review and adjudication.

\noindent\textbf{RQ1 synthesis.} UIPCA helps distinguish a source to recheck, an inference to repair, dependent reasoning to revisit, and a decision to reconsider. Which of these diagnoses is available depends on the interface. An unknown field identifies a limit of the record, not a failure-free part of the run.

\section{What the Evidence Actually Establishes}\label{evidence-mechanisms}

\subsection{Candidates for a Matched Joint Adverse Effect}\label{strict-case-ledger}

Our criterion-level review covers 34 sources identified within the corpus and 24 additional sources retrieved through external follow-up. No reviewed source establishes the matched joint adverse effect specified in Section~\ref{formal-scope-notation} and Equation~\ref{eq:strict-divergence-criterion}. Relaxing only threshold and uncertainty-reporting requirements also yields no positive source. These are source-level judgments, not counts of independent experiments. S3 supplies the selection and sensitivity records; Table~\ref{tab:divergence-evidence} explains the different conclusions supported by candidate findings.

\begin{longtable}[]{@{}
  >{\scriptsize\raggedright\arraybackslash}p{(\linewidth - 8\tabcolsep) * \real{0.1550}}
  >{\scriptsize\raggedright\arraybackslash}p{(\linewidth - 8\tabcolsep) * \real{0.1700}}
  >{\scriptsize\raggedright\arraybackslash}p{(\linewidth - 8\tabcolsep) * \real{0.2050}}
  >{\scriptsize\raggedright\arraybackslash}p{(\linewidth - 8\tabcolsep) * \real{0.2050}}
  >{\scriptsize\raggedright\arraybackslash}p{(\linewidth - 8\tabcolsep) * \real{0.2650}}@{}}
\caption{Representative S3 candidates. \(R\): separately defined reasoning capability; \(F\): factual reliability. Study design and unmet joint-effect criteria are reported separately.}\label{tab:divergence-evidence}\\
\TableTopRule\noalign{}
\rowcolor{TableHeader}\textbf{Comparison} & \textbf{\(R\) Outcome} & \textbf{\(F\) Outcome} & \textbf{Study Design} & \textbf{Unmet Criterion} \\
\TableMidRule\noalign{}
\endfirsthead
\caption[]{(continued)}\\
\TableTopRule\noalign{}
\rowcolor{TableHeader}\textbf{Comparison} & \textbf{\(R\) Outcome} & \textbf{\(F\) Outcome} & \textbf{Study Design} & \textbf{Unmet Criterion} \\
\TableMidRule\noalign{}
\endhead
\TableBottomRule\noalign{}
\endlastfoot
Post-training recipes \citep{al2025are} & No independent \(R\) threshold & Factual QA rises or falls by recipe & Comparison of 12 separately trained systems & Data, reward, and optimizer vary together \TableRowSep
System comparisons \citep{chen2025factuality} & System type, not matched \(R\) & Higher hallucination for two reasoning systems & Reasoning versus non-reasoning systems & Checkpoint and output detail differ \TableRowSep
Natural trace groups \citep{gekhman2026recall} & Recall expansion reported separately & Lower accuracy with an incorrect extracted fact & Observed groups on the same questions & Facts are grouped, not experimentally manipulated \TableRowSep
Qwen3 thinking mode \citep{wang2026margo} & No independent \(R\) & Aggregate gain with some correct-to-incorrect flips & Fixed checkpoint and items; mode varies & Prompts/decoding differ; beneficial flips coexist \TableRowSep
Test-time budget \citep{zhao2025tts} & Tokens, not independent capability & Non-monotonic accuracy and hallucination & Model/task fixed in many budget comparisons & Joint effect unresolved; answer willingness mediates outcomes \TableRowSep
Factuality-aware RL \citep{li2025fspo} & Math reasoning accuracy & Factuality/hallucination measured separately & Training-pipeline comparison & Reasoning and factuality training vary together \TableRowSep
Length-penalized CoT \citep{little2026lengthpenalties} & No-hint accuracy, not independent \(R\) & Monitorability/faithfulness decline; \(F\) unresolved & Length penalty changes the visible trace & Monitorability and \(F\) are different outcomes \TableRowSep
RFC local edits \citep{al2025auditing} & No capability outcome & Accepted edits need not change the final outcome & Within-trajectory intervention & Local acceptance and final reversal use different aggregates \\
\end{longtable}

The candidates concern different claims: pipeline differences, observed reuse, item-level harm alongside aggregate gains, or monitorability rather than factuality. Table~\ref{tab:divergence-evidence} identifies the missing link in each comparison; absence of a strict positive is a design-and-reporting finding, not a field-wide null result.

\subsection{Evidence by Design, Not by Headline}\label{evidence-by-design}

\textbf{Pipeline-confounded comparisons.} Among 12 paired systems, all four cold-start SFT+RL pairs improve on both factual QA sets; the three RL-only pairs decline on TriviaQA, and most RL-only or distillation pairs decline across both tasks \citep{al2025are}. Long-form tests also find more hallucinations for DeepSeek-R1 and QwQ-32B than selected counterparts \citep{chen2025factuality}. These comparisons motivate training hypotheses but do not isolate reasoning: data, rewards, and optimization change together.

A study of post-training separately evaluates access to individual facts and the ability to combine them \citep{yu2026compositioncollapsestablefactual}. Its temporal-chain comparisons are correlational. They support separate reports of factual access and composition, not a universal ranking of training methods.

\textbf{Trace evidence and interventions.} Within one checkpoint, additional computation increases recall, but traces with wrong extracted facts have lower accuracy \citep{gekhman2026recall}. Because the facts were observed rather than inserted, difficulty, confidence, or other trace properties may explain that gap.

Interventions expose hidden detector cues, computation spent on decoys \citep{cheng2025obscures,dnr2025bench}, unavailable tools, and local edits that fail to repair output \citep{al2025how,al2025auditing}. The correction audit is especially informative: accepting a valid edit and recovering the final answer are different outcomes. Its aggregate rates do not establish the probability of recovery after acceptance; S3 preserves their denominators. These tests identify particular mechanisms, not a general trace-length effect.

Audits of visible traces complement these interventions by locating errors: RFC traces repeat unsupported claims \citep{al2025auditing}, graph-coloring traces can invent input edges \citep{heyman2025reasoning}, and RFEval elicits several unfaithful behaviors \citep{han2026rfeval}. Their prevalence estimates depend on the generator, task, and annotation rule.

\textbf{Evaluator and deployment reports.} Grounded-summarization and provider results apply to named systems and dated protocols \citep{vectara2025hallucination,vectara2025nextgen,openai2025personqa}. Vectara's successive suites change documents and evaluation; PersonQA follows its system-card protocol. FaithJudge also shows that few-shot judge choice changes agreement with human labels \citep{tamber2025faithjudge,al2024do}. These scores cannot form one continuous ranking.

\subsection{Five Mechanism Hypotheses}\label{mechanism-hypotheses}

\begin{figure}[t]
  \centering
  \includegraphics[width=\columnwidth]{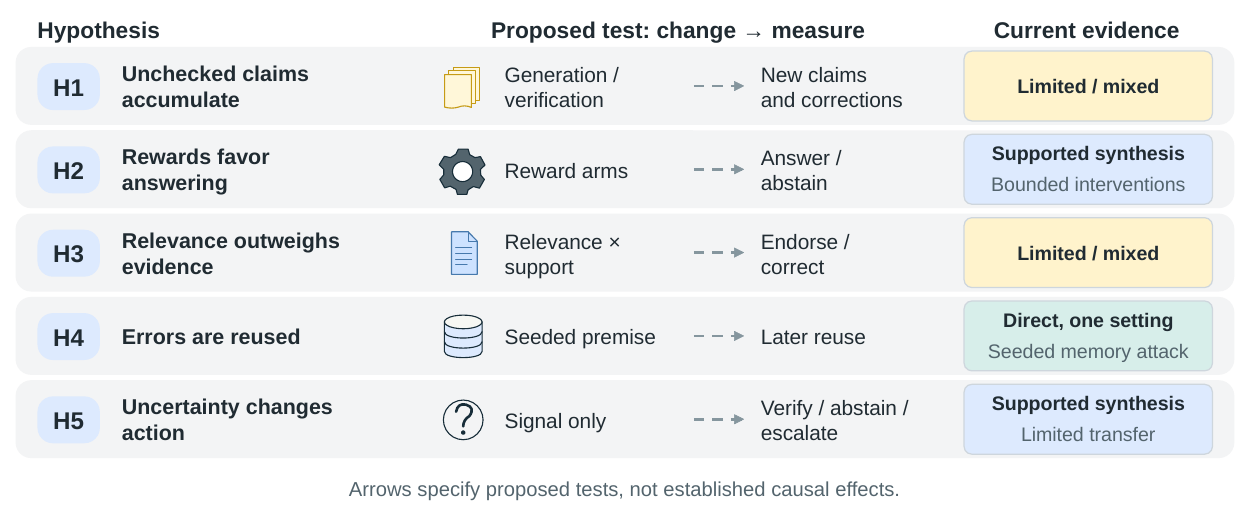}
  \caption{Five hypotheses, their proposed tests, and current evidence. Each row pairs a change with an outcome to measure; dashed arrows do not denote established causal effects. H4 has direct evidence from one seeded memory-attack setting, not an estimate of ordinary error reuse. The tests require the matching controls described in the text. Table~\ref{tab:hypothesis-evidence-grades} specifies the remaining decisive experiments.}\label{fig:mechanisms}
  \Description{Five aligned rows map H1 to generation and verification, H2 to reward arms, H3 to relevance and support, H4 to seeded premises, and H5 to uncertainty signals. Each row shows the outcome to measure and its current evidence grade. H1 and H3 have limited or mixed evidence. H2 and H5 have supported synthesis with bounded scope, while H4 has direct evidence in one seeded memory attack.}
\end{figure}

The evidence grades in Figure~\ref{fig:mechanisms} follow Section~\ref{review-methodology}. Table~\ref{tab:hypothesis-evidence-grades} separates supported links from missing tests; neither single-setting evidence nor synthesis establishes transfer by itself.

\begin{table}[t]
\caption{H1--H5: claims, evidence, and decisive tests. Representative studies support the indicated part of each hypothesis, not every link. The last column gives the remaining experiment.}\label{tab:hypothesis-evidence-grades}
\scriptsize
\renewcommand{\arraystretch}{1.12}
\begin{tabular}{@{}>{\raggedright\arraybackslash}p{0.19\linewidth}>{\raggedright\arraybackslash}p{0.20\linewidth}>{\raggedright\arraybackslash}p{0.24\linewidth}>{\raggedright\arraybackslash}p{0.28\linewidth}@{}}
\toprule
\textbf{Hypothesis} & \textbf{Evidence grade} & \textbf{Representative evidence} & \textbf{Decisive test still missing} \\
\midrule
H1: Unchecked claims accumulate & Limited or mixed & Faulty traces and complexity-related errors \citep{al2025auditing,heyman2025reasoning} & Fix the initial trace; vary unchecked claims, verification, and stopping \\[3pt]
H2: Rewards favor answering & Supported synthesis; bounded reward interventions & Reward granularity and calibrated reflection \citep{palandye2026rewardgranularity,al2025rlcr} & Randomize reward arms from one base checkpoint; fix data, budget, optimization, and evaluation; use multiple seeds \\[3pt]
H3: Relevance outweighs evidence & Limited or mixed & Wrong guidance and sensitivity to supplied evidence \citep{al2025auditing,al2024trusting} & Vary relevance and support for one task using matched wording templates \\[3pt]
H4: Errors are reused & Direct, single-setting seeded attack & FARMA: forged reasoning stored and reintroduced \citep{karamchandani2026farma} & Seed supported, unsupported, and refuted premises; vary independent repair and measure actual reuse \\[3pt]
H5: Uncertainty changes action & Supported synthesis; limited transfer & Uncertainty-guided control and the monitoring--control gap \citep{yan2026upair,yu2026detectingresolvingmonitoringcontrol} & Change only the signal; measure action and useful coverage, then test distribution shift \\
\bottomrule
\end{tabular}
\end{table}

\textbf{H1---unchecked claims accumulate.} Longer traces create more opportunities for error, but also for correction. Faulty traces and complexity-related graph errors fit H1 \citep{al2025auditing,heyman2025reasoning}; improved recall and reduced detector visibility complicate that interpretation \citep{gekhman2026recall,cheng2025obscures,zollicoffer2026sanity}. A test must count corrections as well as new claims while varying verification and stopping. S4 gives a simple independent-error illustration, not an assumption about real traces.

\textbf{H2---rewards favor answering.} Sparse outcome rewards may favor plausible answers over calibrated abstention. Theory and training studies implicate rare facts, noisy gradients, and incorrect reward optima \citep{kalai2026evaluating,li2025fspo,al2025rlvr}. Process and outcome rewards yield different errors in small-model mathematics, under an automated judge \citep{palandye2026rewardgranularity}. Reasoning-finetuned models may also refuse less on unanswerable items or call unavailable tools \citep{song2025tax,kirichenko2025abstentionbench,al2025how}. Randomized reward arms (Table~\ref{tab:hypothesis-evidence-grades}) should measure hallucination, useful coverage, and unnecessary abstention on the same items \citep{al2025rlcr,wu2026mitigating}.

\textbf{H3---relevance outweighs evidence.} Following wrong guidance, agreeing with users, and underusing context are compatible with H3 \citep{al2025auditing,al2024towards,al2024trusting}, but studies operationalize relevance and support differently. Matched wording templates could vary both factors for one task. Preferential endorsement of relevant unsupported claims would support H3; correction in response to contradictory evidence would instead indicate grounding.

\textbf{H4---errors are reused.} Wrong guidance can recur, and traces containing wrong extracted facts have poorer outcomes \citep{al2025auditing,gekhman2026recall}. FARMA directly demonstrates forged reasoning stored in memory and reused \citep{karamchandani2026farma}, establishing an attack path rather than its ordinary frequency. Self-feedback, retained feedback, and tool-assisted checks can also correct errors \citep{madaan2023selfrefine,shinn2023reflexion,al2024critic}. Testing general reuse requires both seeded premise status and independent repair to vary.

\textbf{H5---uncertainty changes action.} Confidence, hidden states, sparse features, and information flow supply different signals \citep{azaria2023the,marks2023the,al2024inside,chen2026hallusae,jia2026informationflow}. Controllers combine uncertainty with progress, critique, or trained reflection \citep{yan2026upair,vijayvargiya2026actionable,huang2026rein}, yet a model may acknowledge contradiction without resolving it \citep{yu2026detectingresolvingmonitoringcontrol}. Activation interventions offer ways to isolate the signal \citep{al2023iti,al2024activation,2026qwenscope}. The decisive outcome is changed verification or abstention at useful coverage, including under distribution shift.

\subsection{Boundary Conditions for Synthesis}\label{synthesis-boundaries}

Four variables explain many apparent disagreements. Output detail changes how many claims can be checked, so long- and short-answer rates are not directly comparable \citep{chen2025factuality,al2023factscore}. Answer willingness changes the answered subset and can produce non-monotonic budget effects \citep{zhao2025tts}. Evaluator access limits available checks, while checkpoint and training changes may dominate the named reasoning mode \citep{al2025are}.

Within-checkpoint recall and between-checkpoint long-form comparisons test different links \citep{gekhman2026recall,chen2025factuality}. Tool interventions identify action failures \citep{al2025how}; observed trace groups suggest intervention targets. Both are informative within scope. Results are excluded when the denominator, source identity, or supported claim cannot be resolved.

Section~\ref{detection-intervention} uses these distinctions to assess when a detection signal can guide verification, repair, or a different release decision.

\noindent\textbf{RQ2 synthesis.} Pipeline differences and harmful traces are documented, but no comparison in the 34-source S3 review or 24 reviewed external sources establishes a matched joint adverse effect. The open question concerns a specified intervention, with reasoning and factuality measured jointly under the controls in Section~\ref{formal-scope-notation}.

\begin{figure}[!t]
  \centering
  \includegraphics[width=\columnwidth]{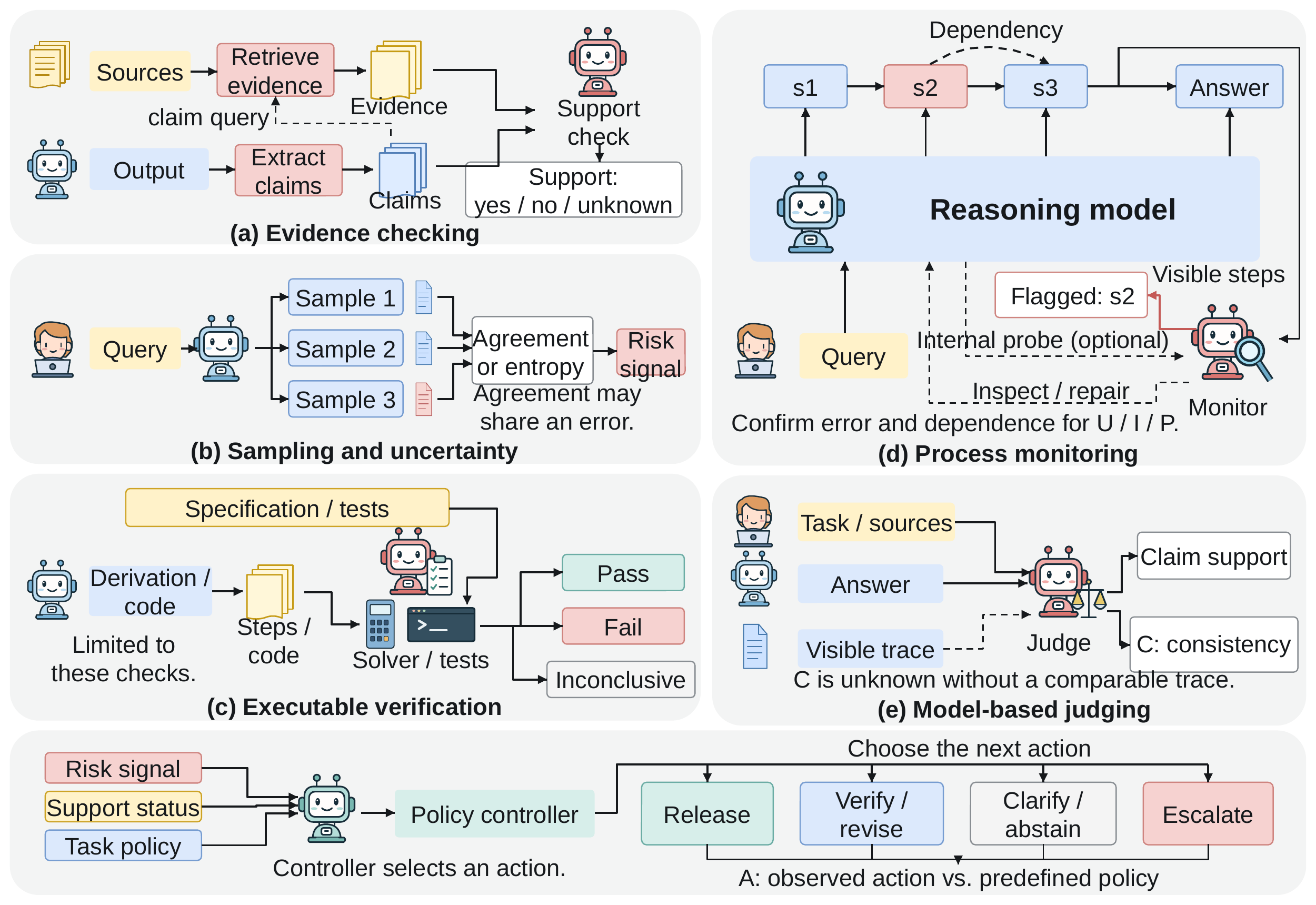}
  \caption{Complementary detection mechanisms and action control. Panels (a)--(e) illustrate evidence checks, sampling, execution, monitoring, and model-based judging. The controller chooses an action; \(A\) audits policy compliance. Table~\ref{tab:detection-action} compares the methods.}\label{fig:detection-landscape}
  \Description{Five panels depict external evidence checking, semantic agreement across samples, executable checks with independent specifications, monitoring of visible steps and optional internal states, and model-based assessment of claim support and visible answer--trace consistency. A bottom strip connects risk signals, evidence status, and task policy to action selection. A separate A audit compares the observed action with the predefined policy.}
\end{figure}

\section{From Detection to Intervention}\label{detection-intervention}

After detecting a suspect claim, the system must decide whether to verify, revise, or withhold the answer. UIPCA identifies the diagnostic target; study design and evidence grades indicate what the supporting studies establish. The method families below connect diagnoses to checks and actions. Interface and model access, timing, and cost determine which interventions are feasible.

\subsection{Signals, Units, and Actions}\label{signals-units-actions}

Output uncertainty, sample agreement, internal monitors, and external evidence serve different purposes. Semantic Entropy and SelfCheckGPT require repeated generations, but agreement is weak evidence when samples share the same false premise \citep{al2024detecting,al2023selfcheckgpt}. Hidden-state and sparse-feature probes can flag cases earlier, although their signals often depend on the checkpoint \citep{al2024semantic,al2026traced,anonymous2025harp,chen2026hallusae}. Claim extraction followed by source checking can provide direct evidence of support, at the cost of extraction, retrieval, and checking \citep{al2023factool,al2024safe,al2024minicheck,al2025clatter,hu2024refchecker}.

Methods also differ in what they check. FActScore evaluates individual facts, while MiniCheck tests claims against documents \citep{al2023factscore,al2024minicheck}. FacTool and SAFE seek external evidence; self-checking instead compares generated outputs \citep{al2023factool,al2024safe,al2023selfcheck,al2024selfcontradictory}. Step detectors target process errors \citep{al2025detection}, and streaming monitors can trigger stopping or repair during generation \citep{al2025auditing,al2026tokenguard}. These units affect the available repair: a sentence score may not locate a faulty step, and an internal anomaly cannot supply a missing fact.

Figure~\ref{fig:detection-landscape} shows checks that can be combined within one system. Icons denote functional roles; sample colors in panel (b) distinguish meanings rather than confirmed errors. Table~\ref{tab:detection-action} groups methods by the decisions they enable. The complete method catalog, source metrics, and locators are in S5.

\begin{longtable}[]{@{}
  >{\scriptsize\raggedright\arraybackslash}p{(\linewidth - 8\tabcolsep) * \real{0.1450}}
  >{\scriptsize\raggedright\arraybackslash}p{(\linewidth - 8\tabcolsep) * \real{0.2100}}
  >{\scriptsize\raggedright\arraybackslash}p{(\linewidth - 8\tabcolsep) * \real{0.1750}}
  >{\scriptsize\raggedright\arraybackslash}p{(\linewidth - 8\tabcolsep) * \real{0.2300}}
  >{\scriptsize\raggedright\arraybackslash}p{(\linewidth - 8\tabcolsep) * \real{0.2400}}@{}}
\caption{Detection-to-action comparison. Component mappings identify possible uses of a signal, not a claim that every listed method directly annotates UIPCA.}\label{tab:detection-action}\\
\TableTopRule\noalign{}
\rowcolor{TableHeader}\textbf{Signal / Examples} & \textbf{Unit / Covered Component} & \textbf{Interface / Access} & \textbf{Feasible Action} & \textbf{Boundary and Cost} \\
\TableMidRule\noalign{}
\endfirsthead
\caption[]{(continued)}\\
\TableTopRule\noalign{}
\rowcolor{TableHeader}\textbf{Signal / Examples} & \textbf{Unit / Covered Component} & \textbf{Interface / Access} & \textbf{Feasible Action} & \textbf{Boundary and Cost} \\
\TableMidRule\noalign{}
\endhead
\TableBottomRule\noalign{}
\endlastfoot
External evidence: CLATTER, RefChecker \citep{al2025clatter,hu2024refchecker} & Claim; U and source-backed I & All interfaces; source required & Accept, retrieve, revise, abstain & Strong support; extraction and checking cost grows with the number of claims \TableRowSep
Uncertainty or learned scoring: Semantic Entropy Probes, FaithLens, UPAIR \citep{al2024semantic,al2025faithlens,yan2026upair} & Answer/trace; routes checks rather than establishing support or C & White-box probe or learned detector & Verify, resample, reject & Checkpoint and training-domain dependence \TableRowSep
Verification and calibrated action: VeriFY, CURE, Latent Critic \citep{al2025verify,liu2026cure,vijayvargiya2026actionable} & Claim/answer; support checks and action selection & Black-box model, labels, or internal critic & Answer, verify, abstain & Coverage depends on boundary estimation and detector transfer \TableRowSep
White-box trajectory: TRACED, HARP, Reasoning Denoiser \citep{al2026traced,anonymous2025harp,fang2026reasoningdenoiser} & Step/trajectory; I and P routing & Internal states or trained features & Stop, inspect, escalate & Early but checkpoint-specific; no external factual evidence \TableRowSep
Online search/repair: HaluSearch, Token-Guard \citep{cheng2025think,al2026tokenguard} & Prefix/segment; I and P & Generation-time score and access & Redirect, prune, regenerate & Detection and corrective policy effects are coupled \TableRowSep
Executable and grounded checks \citep{ling2023deductive,pan2023logiclm,ma2026hkvlm,jin2026refact} & Constraint, region, program state; U/I & Evidence, tools, or sandbox & Bind, execute, repair, abstain & The check covers only declared evidence and tested properties \TableRowSep
Human or expert review & Consequential claim/decision; all observable components & Structured case packet & Clarify, revise, escalate & Often slow and costly; quality depends on evidence presentation \\
\end{longtable}

A fast monitor is most useful when it identifies cases that benefit from a stronger check. Its ranking accuracy alone does not show whether those checks prevent wrong answers or create excessive review work. Detector comparisons and complete-system comparisons therefore answer different questions; S5 gives the conditions needed for each.

Error location guides repair. Case A needs an arithmetic check before reuse; Case B needs a constraint check at Step 8, although an answer--trace comparison can still detect \(C\). PRISM likewise separates memory, instruction, and reasoning failures for diagnosis \citep{wu2026prism}. For control, the value of localization lies in improving retrieval, repair, or escalation.

A trace can guide computation without faithfully explaining it: chemistry drafts remain useful even when their verbal claims lack support \citep{li2026chemicalcot}. Conversely, reasoning text may persuade a monitor or lose warning cues after length-penalized training \citep{za2026persuasion,little2026lengthpenalties}. A routing signal is therefore not itself evidence for a claim.

Access further limits these uses. A provider may run an internal monitor that a user cannot inspect or reproduce. A user-facing summary supports checks on disclosed claims only; omitted process evidence remains \(\bot\), meaning insufficient information.

\subsection{Intervention Families}\label{intervention-families}

Inference-time methods add computation, evidence, or earlier verification. Budget and length controls alter latency \citep{anthropic2025claude,al2025overthinking}. Extra deliberation can improve recall, but its effects on factuality and willingness to answer vary across models \citep{gekhman2026recall,zhao2025tts,jung2026phantombench}. Multi-agent debate adds paths that may share source errors \citep{du2023multiagent}. Answer rate and selective accuracy distinguish reliability gains from increased refusal. Completion, path diversity, and cost help assess extra computation; trace length alone establishes neither repair nor waste.

Independent evidence gives a disputed claim a stronger check than reconsideration alone. Verification, retrieval, and tool-assisted critique can offer such checks \citep{ling2023deductive,al2020retrievalaugmented,yao2023react,al2024critic,al2025rrag}, although weak retrieval or unavailable tools can introduce errors \citep{al2025how}. Comparing prevention, interruption, rejection, and recovery therefore requires matched evidence and cost budgets.

Early verification can reject a faulty step before reuse. Decoding-time mode switches or verification can interrupt generation \citep{al2025dsvd}, while HaluSearch couples detection with search \citep{cheng2025think}. Clinical critique-and-regenerate methods instead operate on completed answers \citep{liao2026medheal}. They can improve what is released while leaving the original trace unrepaired. Both approaches require an independent final check.

Process supervision moves checking into training. Instead of rewarding only the final answer, it provides feedback on intermediate steps. Labels backed by source evidence can identify unsupported premises (\(U\)); rule or execution checks can identify invalid transitions (\(I\)) \citep{al2021training,al2023lets}. Detecting propagation (\(P\)) requires more: a later step must be shown to depend on a confirmed error. Knowing that both steps are wrong does not establish that connection.

Process/outcome reward comparisons in mathematics and step verification for scientific tools illustrate these differences \citep{palandye2026rewardgranularity,zhao2026sciprm}. Factuality and faithfulness objectives target other units \citep{li2025fspo,nie2026faithrl,al2025knowrl}. A step reward can miss a false premise; a factuality verifier can miss an invalid transition. Table~\ref{tab:verifier-lifecycle-map} relates supervision to these faults.

\begin{table}[H]
\caption{Process supervision and verifier placement by lifecycle target. No verifier type covers all components.}\label{tab:verifier-lifecycle-map}
\scriptsize
\begin{tabular}{@{}p{0.24\linewidth}p{0.09\linewidth}p{0.25\linewidth}p{0.33\linewidth}@{}}
\toprule
\textbf{Supervision or verifier} & \textbf{Target} & \textbf{Useful intervention} & \textbf{Characteristic failure} \\
\midrule
Evidence-backed premise check & \(U\) & Reject or retrieve before a premise enters & Open-world evidence gaps; judge shares generator bias \\
Rule, math, or execution check & \(I\) & Reject an invalid local transition & Incomplete checker; valid alternative paths scored as wrong \\
Dependency-aware reward & \(P\) & Stop reuse or invalidate a branch & Missing edges; rewards favor style or late correction rather than preventing reuse \\
Answer--trace or action verifier & \(C/A\) & Compare the answer with the trace; revise, verify, abstain, or escalate & Late detection; exploitable scores, unnecessary refusal, or an unrepaired trace \\
\bottomrule
\end{tabular}
\end{table}

Final-outcome rewards can improve answers while leaving \(U/I/P\) unmeasured. Establishing a change in answer--trace consistency (\(C\)) also requires a trace check. Confidence and abstention objectives instead train uncertainty expression or action choice \citep{liu2026cure,wu2026mitigating,al2025verify,zhai2026abstainr1}. Long-form factuality RL balances precision, detail, and relevance \citep{chen2025factuality}.

The feedback itself can also be unreliable. Claim-level falsification concentrates checks on decision-relevant content, so its value depends on extraction and verifier quality \citep{xu2026claimlevel}. Automated labels may inherit judge bias or reject valid alternatives; optimization may exploit the score or favor preferred wording. Evaluating these gains therefore requires independent evidence and an audited subset.

An overall score can also hide differences between outcomes. Reasoning training may amplify deliberative language without correctness-associated behaviors \citep{nyandwi2026amplified}. Changing the reward alters error patterns \citep{palandye2026rewardgranularity}, while penalizing length reduces monitoring cues \citep{little2026lengthpenalties}. Autoformalizers illustrate a related distinction: a proof may concern a silently repaired statement rather than the original one \citep{cornish2026faithformbench}. Separate checks on reasoning, monitoring, and fidelity to the task reveal what the aggregate gain leaves unresolved.

Attribution to a particular verifier poses a separate question. When data, supervision, and optimization change together, any overall gain concerns the full pipeline (Section~\ref{formal-scope-notation}). For deployment, the decisive question is whether the improved check leads to better subsequent answers or actions without excessive refusal or cost.

Other interventions change generation more directly. Decoding methods adjust scores for candidate tokens, while activation methods steer internal representations using signals associated with truthfulness \citep{al2023dola,al2023iti}. Knowledge editors such as ROME and MEMIT modify factual associations in model parameters \citep{al2022rome,al2023memit}. Multi-hop and sequential tests reveal that an edited fact may not carry through to dependent answers, and unrelated knowledge may also change \citep{al2023mquake,al2024model}. The edited answer is therefore only one part of the outcome to assess.

Across domains, the same diagnostic questions call for different evidence and checkers. SPD-Faith separates missed perception from unused evidence; HKVLM checks region binding; Reflect-R1 retrieves video evidence \citep{2026spdfaith,ma2026hkvlm,chen2026reflectr1}. Joint changes to representation and generation require separate grounding, reasoning, and answer scores \citep{al2024vcd,al2024opera,al2022learn}. CodeHalu, EvalPlus, and SWE-bench test execution or repair at different units \citep{tian2024codehalu,liu2023evalplus,jimenez2024swebench}. Passing selected tests establishes neither security nor satisfaction of all requirements.

\subsection{Worked Interface-to-Action Comparison}\label{worked-control-comparison}

Consider three views of Case A: the full trace, an illustrative summary, and answer only. All show remainder 27 for \(1024\times1047\times1050\times1053\) divided by 33, instead of 3. They are views of one completed run, not additional model runs or observed provider summaries.

The full trace exposes \(1024\bmod33=29\) at Step 3; recalculation gives 1, confirming \(I_3=1\). Later reuse establishes \(P\). Repair must recheck the other residues and dependent products. Agreement with the wrong trace gives \(C=0\), not arithmetic correctness.

A summary that mentions modular arithmetic and remainder 27 omits the reductions needed to identify \(I/P\) and \(\tau^\star\). With answer only, \(C\) is also unavailable. Yet independent calculation from the question gives 3 under either restricted view: the result remains checkable even when its origin is hidden.

Under a hypothetical policy requiring independent numerical verification, all views would block 27 and trigger correction; incomplete verification would require abstention. The source case still has \(A=\bot\), since it declares no such policy.

Thus localization and answer verification offer distinct benefits. A full trace can reveal where to restart the calculation; external checking can still prevent the wrong answer from being released when the trace is hidden. Whether the additional localization is worth its cost is an empirical question, not an automatic consequence of greater visibility.

\subsection{A Calibrated Control Cascade}\label{control-cascade}

Hallucination control begins by splitting the answer and visible reasoning into claims, then checking each claim against declared evidence. Claims are supported, unsupported, or unknown. When visible information permits, unsupported claims are traced to their origin and receive component-specific repair. Unknown claims go to retrieval, clarification, abstention, or escalation; they never count as supported. Harm, reversibility, coverage, and cost guide action selection. The subsequent \(A\) audit checks policy compliance without changing evidence status.

\begin{figure}[t]
  \centering
  \includegraphics[width=\columnwidth]{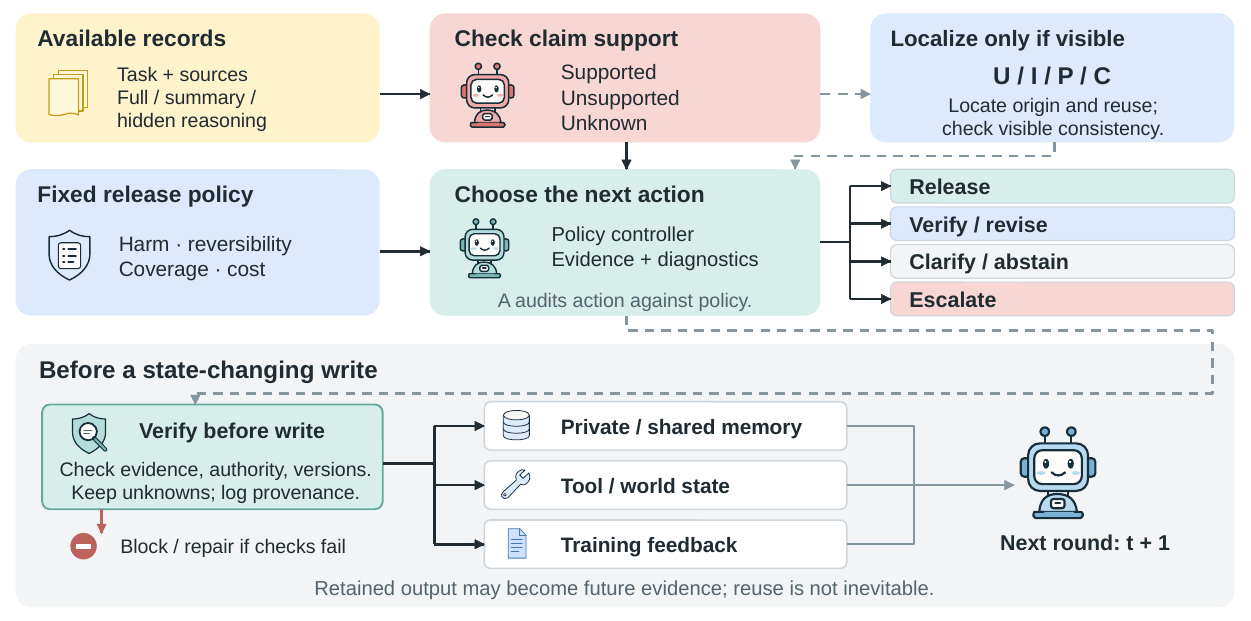}
  \caption{From claim checks to release and persistent-state control. Claim support and visible-process diagnostics inform a controller governed by a fixed policy. Harm, reversibility, coverage, and cost affect action selection, not evidence status. \(A\) audits the chosen action against that policy. State-changing operations also require evidence, write-authority, and version checks. Private or shared memory, tool or world state, and training feedback are parallel destinations. Dashed paths are conditional; retained output may become evidence in a later round, but reuse is not inevitable.}\label{fig:deployment-workflow}
  \Description{Available records feed claim checking and, when visible information permits, process localization. Evidence and a separate fixed policy feed action selection: release, verify or revise, clarify or abstain, or escalate. Before any state-changing operation, a write gate binds evidence, preserves unknowns, checks write authority and versions, and logs provenance and repairs. Authorized writes may enter private or shared memory, tool or world state, or training feedback in parallel. Failed checks lead to blocking or repair; retained state may become input to a later round.}
\end{figure}

The practical advantage of a cascade is that expensive checks need not run on every claim. A fast screen can select cases for retrieval, execution, or expert review, while the task determines which check can resolve the doubt. A factual dispute may need a source; an arithmetic dispute may need recalculation rather than more generated explanation.

The stages nevertheless cannot be judged in isolation. A screen changes which cases reach the verifier, and a conservative controller can appear reliable simply by escalating almost everything. Two quantities make this trade-off visible: \emph{coverage} is the fraction of requests answered, and \emph{selective risk} is the error rate among those answers. The complete system must be assessed alongside its repair success, workload, and cost. S5 gives the reporting protocol.

\noindent\textbf{RQ3 synthesis.} No detector family covers every evidence object, interface, and action. Fast signals can route cases to stronger evidence checks, but their value depends on transfer, false positives, and downstream workload. Consequential release decisions require evaluation of the complete routing and verification process.

\section{Evaluation, Deployment, and Research Agenda}\label{evaluation-deployment-agenda}

\subsection{Benchmark Coverage and Joint Metrics}\label{benchmark-coverage}

Benchmark families answer different questions. QA and grounding suites assess outcomes \citep{wei2024simpleqa,al2017triviaqa,al2022truthfulqa,al2023factscore,al2024minicheck}; process suites expose or perturb steps \citep{al2025faithcotbench,mittal2026c2faith,al2025race}; selective suites test confidence or answerability \citep{al2026rmcb,kirichenko2025abstentionbench,jung2026phantombench}. Domain suites add perception or execution \citep{al2024hallusionbench,al2025rhbench,tian2024codehalu,gu2024cruxeval}. Table~\ref{tab:benchmark-reporting} compares their direct evidence and missing companion measurements.

The choice of benchmark determines which diagnosis is possible. An answer score cannot locate the first reasoning error without a trace. Conversely, a step-labeled dataset may reveal an invalid inference without showing whether the system should have answered or abstained. These are complementary measurements, not interchangeable reliability scores.

Domain scores remain conditional. Objects, grounding, and answers do not reveal whether a visual region was used \citep{al2023pope,al2024hallusionbench,al2024amber}. State prediction, tests, and repair do not establish full requirements or security \citep{tian2024codehalu,gu2024cruxeval,liu2023evalplus,jimenez2024swebench}. Long-context and dialogue tasks additionally need retrieval and binding records.

Memory evaluation extends this comparison beyond the current context. LoCoMo tests long-term conversational recall \citep{maharana2024locomo}, while LongMemEval includes knowledge updates and abstention \citep{wu2025longmemeval}. MemoryAgentBench adds incremental learning and selective forgetting \citep{hu2025memoryagentbench}. LongMemEval-V2 tests evidence gathered from past web-agent trajectories, including false-premise questions \citep{wu2026longmemevalv2}. These tasks assess what memory makes available to a later answer.

HaluMem instead scores extraction, updates, and memory-based QA separately, providing closer access to the operation where content becomes unsupported \citep{chen2025halumem}. MemoryData evaluates component choices, retrieval fidelity, update behavior, and cost \citep{zhou2026memorydata}. Neither a memory-operation label nor a retrieved source ID by itself establishes dependent reuse under UIPCA.

\begin{longtable}[]{@{}
  >{\scriptsize\raggedright\arraybackslash}p{(\linewidth - 6\tabcolsep) * \real{0.2250}}
  >{\scriptsize\raggedright\arraybackslash}p{(\linewidth - 6\tabcolsep) * \real{0.1800}}
  >{\scriptsize\raggedright\arraybackslash}p{(\linewidth - 6\tabcolsep) * \real{0.2850}}
  >{\scriptsize\raggedright\arraybackslash}p{(\linewidth - 6\tabcolsep) * \real{0.3100}}@{}}
\caption{Benchmark families and reporting coverage needed for a release decision. Named resources illustrate each family; protocols and metrics are not interchangeable. Memory governance is separated from answer quality and semantic memory errors.}\label{tab:benchmark-reporting}\\
\TableTopRule\noalign{}
\rowcolor{TableHeader}\textbf{Family} & \textbf{Primary Unit} & \textbf{Directly Reveals} & \textbf{Required Companion Report} \\
\TableMidRule\noalign{}
\endfirsthead
\caption[]{(continued)}\\
\TableTopRule\noalign{}
\rowcolor{TableHeader}\textbf{Family} & \textbf{Primary Unit} & \textbf{Directly Reveals} & \textbf{Required Companion Report} \\
\TableMidRule\noalign{}
\endhead
\TableBottomRule\noalign{}
\endlastfoot
Factual QA / grounding: SimpleQA, TriviaQA, FACTS \citep{wei2024simpleqa,al2017triviaqa,al2025the} & Answer or claim--source pair & Outcome factuality or source support & Evidence/evaluator version; extraction rule; answer rate and coverage \TableRowSep
Trace / perturbation: FINE-CoT, C2-Faith, RACE \citep{al2025faithcotbench,mittal2026c2faith,al2025race} & Step, dependency, trace, answer & Selected process errors; coverage depends on suite labels & Segmentation, codebook, boundary cases, interface, unobserved steps \TableRowSep
Calibration / answerability: RMCB, VeriFY, AbstentionBench \citep{al2026rmcb,al2025verify,kirichenko2025abstentionbench} & Confidence, answer, action & Ranking, calibration, answer/abstain behavior & Brier/ECE, selective risk, coverage, unnecessary refusal \TableRowSep
Long context / changing evidence: RULER, Lost in the Middle, StreamingQA, HalluDial \citep{al2024ruler,al2024lost,al2022streamingqa,al2024halludial} & Retrieved answer or turn & Outcome under available, positioned, or changing evidence & Retrieval/binding record, evidence time, first-error annotation \TableRowSep
Multimodal: POPE, HallusionBench, AMBER \citep{al2023pope,al2024hallusionbench,al2024amber} & Object, region, claim, answer & Perception/grounding and selected outcome & Coverage, binding, reasoning, answer, abstention as separate fields \TableRowSep
Code / software: CodeHalu, CRUXEval, EvalPlus, SWE-bench \citep{tian2024codehalu,gu2024cruxeval,liu2023evalplus,jimenez2024swebench} & Program state, function, repair, patch & Selected execution, repair, or issue-resolution behavior & Environment, tests/coverage, analyzer, security, repository snapshot \TableRowSep
Long-term memory: LoCoMo, LongMemEval, MemoryAgentBench, LongMemEval-V2 \citep{maharana2024locomo,wu2025longmemeval,hu2025memoryagentbench,wu2026longmemevalv2} & History, retrieved context, answer & Recall, updates, forgetting, or environment knowledge; some suites include abstention & History version, evidence-bearing turns, retrieval budget, supported answers and unnecessary refusal \TableRowSep
Memory operations: HaluMem, MemoryData \citep{chen2025halumem,zhou2026memorydata} & Extracted item, memory update, retrieval result & Semantic memory errors or component-level quality and cost & Before/after state, source support, evaluator, operation failures, downstream outcomes \TableRowSep
Shared-memory governance: ArgusFleet \citep{margalit2026governed} & Authorized write/read, supersession, provenance chain & Scope enforcement, state visibility, and conflict handling in the tested service & Identity/scope, versions, concurrency, read path, dependent use, revocation and recovery \\
\end{longtable}

\subsection{Critical Benchmark Gaps}\label{critical-benchmark-gaps}

Three weaknesses recur across the benchmark catalog. \textbf{First, outcome-only scoring hides the path.} One score can collapse origin, reuse, answer--trace consistency, and action. It cannot show whether a correct answer relied on an unsupported trace or where a wrong answer began.

\textbf{Second, many labels depend on an LLM judge.} Judge setup can change agreement with audited labels, and surface cues may replace the intended factuality test \citep{tamber2025faithjudge,al2024do}. FaithformBench tests whether controlled valid and invalid inputs preserve their status \citep{cornish2026faithformbench}. HALLMARK shows that false positives can dominate verifier deployment even when headline recall is high \citep{reizinger2026hallmark}.

Errors may be correlated when generator and judge share a model family or training style. Their agreement can then reflect a shared mistake rather than independent confirmation. A judge is useful for routing review, but its reliability has to be checked against evidence outside that agreement.

Memory benchmarks make this separation especially important. HaluMem uses GPT-4o for both memory-based answer generation and scoring, although dataset construction includes human quality checks \citep{chen2025halumem}. Agreement within that pipeline is not an independent factuality guarantee. Judge updates create another ambiguity: a changed score may reflect a changed evaluator rather than better memory. Stable reference cases and independent evidence are needed to distinguish the two.

\textbf{Third, action coverage is thin.} Methods increasingly use signals to answer, stop, revise, or abstain. A method demonstration is not a benchmark for the full policy. Existing tasks rarely cover clarification, retrieval, tool verification, expert escalation, memory-write denial, or rollback. Shared benchmarks provide the least coverage of action-policy evaluation (\(A\)). Better benchmarks need process labels, externally checked judges, and complete action and cost records.

\subsection{A Lifecycle-Aware Construction Agenda}\label{benchmark-construction-agenda}

The missing benchmark is not simply a larger collection of wrong answers. It is a collection that connects an error, the evidence available to detect it, and the action taken in response. Different views of the same run can reveal how much diagnosis is lost when steps are summarized or hidden. Controlled changes to premises or inferences can then test whether a proposed safeguard prevents harm, rather than merely recognizing difficult cases.

Evaluation must also distinguish better answers from fewer answers \citep{al2025rlcr,al2026rmcb,liu2026cure}. Confidence metrics address only part of that distinction. AUROC measures how well a score ranks cases, not whether its numerical confidence is accurate. Brier score and expected calibration error (ECE) assess probability predictions in different ways; S4 gives their definitions and limitations. Similarly, evidence that a source supports a step is not evidence that the step influenced the answer \citep{al2025clatter,al2025race,al2026traced}.

VeriFY illustrates why metric definitions matter. Its model-relative precision is correctness among released answers; recall compares correct answers with correct answers plus unnecessary abstentions \citep{al2025verify}. Recall thus depends on how the protocol establishes the base model's knowledge, not objective answerability alone. S4 gives the count definitions.

For an actionable benchmark, these measurements need to accompany the same cases. A useful error location should help repair an answer or prevent an unsafe action, not merely improve a step-ranking score. S5 supplies the detailed construction and reporting protocol, including escalation, write denial, and rollback.

\subsection{Interface-Aware Release Policy}\label{release-policy}

A release policy decides whether to answer, gather evidence, revise, ask for clarification, abstain, or escalate. It must balance avoiding errors with providing useful answers. Refusing everything avoids releasing a wrong answer but provides no useful coverage; answering everything may exceed the application's tolerance for harm. The policy's operating point depends on that tolerance and the capacity for further review.

For illustration, releasing 6,000 of 10,000 requests with 30 wrong answers gives an observed selective risk of \(0.5\%\) and coverage of \(60\%\). These estimates meet a maximum risk of \(0.6\%\) and minimum coverage of \(55\%\), but not a maximum risk of \(0.2\%\). Sampling uncertainty and changes in operational data still matter. The thresholds should reflect tolerated harm and useful service, not be chosen after evaluation to make a policy appear successful.

Interface visibility limits which evidence can inform that choice. Full traces allow process checks; summaries allow checks only on disclosed claims. An unavailable process check cannot be assigned zero error. If the risk and coverage requirements cannot both be met with the available evidence and resources, the policy is infeasible. S4 gives the optimization template; S5 gives threshold-selection and uncertainty procedures.

The same signal can justify different actions. A reversible writing error may warrant a warning; a dose recommendation missing patient information may require clarification or clinical review \citep{liao2026medheal,kirichenko2025abstentionbench}. Passing software tests may still leave security unchecked \citep{liu2023evalplus,jimenez2024swebench}. Score the action taken: expressed uncertainty is not protection unless it changes unsafe release.

Operational comparisons can otherwise reward extra refusal or unreported computation rather than better control \citep{al2024safe,hu2024refchecker,al2025overthinking,al2026tokenguard}. The relevant outcome is what the complete system releases or does, including the cost of obtaining its evidence.

\subsection{From Released Answers to Persistent State}\label{persistent-state}

An answer can outlive its turn through memory, shared state, tool effects, or training feedback. Text revision, action rollback, and parameter correction are different recovery operations. Each raises the same question: what evidence is required before a claim persists beyond the current turn?

The return path in Figure~\ref{fig:deployment-workflow} requires reuse before repair, not merely retention. Self-Refine supplies self-feedback, Reflexion retains feedback across trials, and CRITIC adds tool-assisted checks \citep{madaan2023selfrefine,shinn2023reflexion,al2024critic}; iteration alone establishes no repair. FARMA demonstrates the opposite path: forged reasoning histories are stored and reintroduced \citep{karamchandani2026farma}. This is the single-setting attack evidence for H4.

Existing memory controls act at different points (Table~\ref{tab:memory-controls}). A-MAC decides which candidate memories to admit, update, or reject using a weighted admission score \citep{zhang2026amac}. Its factual-confidence feature is ROUGE-L overlap with supporting conversation spans, so lexical similarity can influence admission without establishing entailment. TrustMem instead evaluates the transition from old to new memory: a frozen LLM checks coverage, preservation, and support, and its scores guide memory-policy training \citep{yang2026trustmem}. This supplies update-level feedback, but support inherited from old memory still needs scrutiny if that memory is already wrong. Admission and consolidation therefore require different checks: whether an item should enter, and whether an edit preserves valid information without adding unsupported claims.

Shared memory adds access and consistency requirements. Governed Memory combines typed properties, entity-scoped retrieval, and organizational policy routing; its reported conflict tests concern sequential writes, not concurrent writers \citep{taheri2026governedmemory}. MemClaw, evaluated through ArgusFleet, attaches scope, supersession, and provenance to shared records \citep{margalit2026governed}. Its evaluation reports a concrete interaction between safeguards: a synchronous near-duplicate gate can reject a contradictory write before the asynchronous conflict detector receives it. A gate intended to prevent redundancy can thus obstruct a legitimate correction. This is evidence about control ordering in one service, not a general rate of memory hallucination.

The distinction changes what an \(A\) audit must inspect. With private memory, the record can focus on one agent's write decision and its required checks. With shared memory, it must also identify the writer's authority, the versions visible to readers, and the scope of correction or revocation. The definition of \(A\) remains a violation of a fixed policy; the policy now governs several actors. Provenance can show who wrote or derived a record, but it does not establish that the record is true. Nor does visibility to another agent establish \(P\): the later claim must depend on a confirmed error. A valid, unauthorized disclosure is a governance failure even when no hallucination occurs.

These distinctions also clarify recovery. Denying a write prevents one form of persistence; superseding a record changes which version should be retrieved. Neither operation alone repairs summaries, tool actions, or feedback already derived from that record. A recovery evaluation should therefore follow affected consumers after correction, not stop when the storage API acknowledges an update. The memory benchmarks in Table~\ref{tab:benchmark-reporting} supply parts of this evidence, while a joint test of semantic correction, authorized access, and downstream recovery remains necessary.

\begin{table}[!t]
\caption{Persistent-memory controls and the evidence they provide. Systems are compared by control point, not by scores from different tasks. Evaluation resources in Table~\ref{tab:benchmark-reporting} serve a separate role.}\label{tab:memory-controls}
\centering
\footnotesize\setlength{\tabcolsep}{3pt}\setlength{\extrarowheight}{0.25pt}\renewcommand{\arraystretch}{1.00}
\begin{tabular}{@{}
  >{\scriptsize\raggedright\arraybackslash}p{(\linewidth - 6\tabcolsep) * \real{0.1800}}
  >{\scriptsize\raggedright\arraybackslash}p{(\linewidth - 6\tabcolsep) * \real{0.2500}}
  >{\scriptsize\raggedright\arraybackslash}p{(\linewidth - 6\tabcolsep) * \real{0.2700}}
  >{\scriptsize\raggedright\arraybackslash}p{(\linewidth - 6\tabcolsep) * \real{0.3000}}@{}}
\TableTopRule\noalign{}
\rowcolor{TableHeader}\textbf{System} & \textbf{Control point} & \textbf{Check or evidence} & \textbf{Remaining boundary} \\
\TableMidRule\noalign{}
A-MAC \citep{zhang2026amac} & Candidate admission and conflict-aware replacement & Weighted utility, overlap-based confidence, novelty, recency, and type prior & Admission precision/recall does not establish semantic truth; overlap is a support proxy \TableRowSep
TrustMem \citep{yang2026trustmem} & Learned write, revise, and prune behavior & Transition verifier scores coverage, preservation, and faithfulness for reward and preference training & Frozen-LLM judgments depend on the supplied chunk and prior memory; no independent truth guarantee \TableRowSep
Governed Memory \citep{taheri2026governedmemory} & Typed extraction, entity-scoped retrieval, policy routing & Heuristic quality gates, schema checks, provenance metadata, and bounded retrieval & Self-evaluation; sequential conflict tests leave concurrent multi-agent writes unresolved \TableRowSep
MemClaw / ArgusFleet \citep{margalit2026governed} & Scoped access, supersession, and controlled propagation & Service traces test authorized visibility, conflict handling, and provenance reconstruction & One-service self-evaluation; admission can pre-empt correction; visibility is not dependent reasoning use \\
\TableBottomRule\noalign{}
\end{tabular}
\end{table}

The useful quantity here is not how often a claim is stored, but how many later consumers reuse it before independent repair. We denote that expected count by \(\mathcal R_{H,t}\). A consumer may be another agent, a later reasoning round, or another operation that uses the retained claim. Its identity and the observation horizon must be fixed for the count to be meaningful.

For a simple illustration, suppose three consumers each have a \(50\%\) chance of selecting a claim, and a selected claim has a \(20\%\) chance of repair before use. The expected number of unrepaired uses is then \(3\times0.5\times0.8=1.2\). Raising the conditional repair probability to \(60\%\) reduces it to \(0.6\). These are hypothetical values, not observed rates. S4 gives the general definition, which also accommodates different reuse and repair probabilities across consumers. S5 describes how to compare checks before writing with correction after reuse.

\subsection{Experiments That Close the Open Links}\label{research-agenda}

Three open questions follow from this synthesis. The first concerns the apparent trade-off between reasoning and reliability. Extra computation may help when it includes verification, while abstention rewards may improve reported accuracy chiefly by reducing coverage. The matched designs in Section~\ref{formal-scope-notation} can separate these explanations. Their purpose is to identify which intervention changes which outcome, not to rank ``more reasoning'' as universally better or worse.

The second concerns the practical value of localization. Does knowing the error's location improve repair beyond a response-level warning? Comparing these signals on the same cases would test UIPCA's proposed diagnostic benefit more directly than agreement with a label alone. Restricted interfaces are important here: a useful system must act on the evidence it can access, without filling hidden fields with inferred certainty.

The third concerns transfer and recovery. A safeguard can succeed on one task because it has access to decisive evidence that is unavailable elsewhere. Translation may change entities, negation, units, or evidential strength \citep{al2023hallucinations,team2022scaling}; context position and visual grounding can change which evidence is used \citep{al2024longbench,al2024lost,al2025rhbench,2026spdfaith}. In code, a passing result remains conditional on requirements and tests \citep{erfanian2026delulu,gu2024cruxeval,liu2023evalplus,jimenez2024swebench}. Transfer therefore concerns the adequacy of the check, not just the model's score in a new domain.

Recovery also depends on timing. Correcting a premise before reuse and repairing a final answer after reuse solve different problems. Local edit acceptance may leave the outcome unchanged \citep{al2025auditing}, and retrieval or tools may move the first error rather than remove it \citep{al2025rrag,al2025how}. Comparing early checks, branch interruption, and later correction under common conditions would show where a safeguard still prevents harm. Domain shifts, adversarial evidence, and unavailable tools test whether that benefit survives beyond the original setting \citep{li2025fspo,al2025verify,liu2026cure,wu2026mitigating}. S3 and S5 preserve the experimental controls and statistical procedures for these questions.

\noindent\textbf{RQ4 synthesis.} Answer quality, visible process evidence, coverage, action, and cost require separate measurements. Shared benchmarks still cover few clarification, escalation, persistent-write, and rollback decisions. Policy evaluation must therefore specify a risk--coverage operating point and extend beyond the displayed answer to consequential writes.

\section{Conclusion}\label{conclusion}

UIPCA separates unsupported premises and invalid inferences, dependent reuse, visible answer--trace consistency, and action-policy failure. These are different diagnostic questions, not sequential stages. Text-centered evidence and selected transfer cases show why repair should follow the earliest useful observable check. Hidden process evidence remains unknown; the record does not infer deceptive intent.

Pipeline differences and harmful traces do not establish that reasoning generally lowers factual reliability. No source in the 34-entry S3 review or the 24-source external follow-up meets the specified matched joint adverse-effect criterion. This bounded finding leaves the causal question open. H1--H5 identify tests that separate the unresolved links while measuring action, coverage, and cost.

Answer factuality, process support, confidence, and action need separate measurements. Their joint evaluation distinguishes better evidence from increased refusal or computation. Consequential outputs also require checks before they enter memory, shared state, tools, or training feedback. Verification and provenance can support writes; unresolved content may need quarantine, and harmful writes may need rollback. Reliability concerns both the current answer and the state inherited by later rounds.

\renewcommand{\showeprint}[2][]{}
\bibliographystyle{ACM-Reference-Format}
\bibliography{references}

\end{document}